%% file: mobicom.tex
\documentclass[sigconf, 10pt]{acmart}

\usepackage{booktabs} % For formal tables
\usepackage{algorithm}
\usepackage{microtype}
\usepackage{setspace}
\usepackage{marvosym}
\usepackage{hyperref}

\usepackage{algpseudocode}

\usepackage[caption=false,subrefformat=parens]{subfig}
\usepackage{xspace}

\usepackage{amssymb,amsmath}
\usepackage{color}
\usepackage{balance}
\usepackage{soul}
\usepackage{cancel}
\usepackage{threeparttable}
\usepackage[normalem]{ulem}
\usepackage{makecell}
\usepackage{tabularx}
\usepackage{wrapfig,adjustbox}
\usepackage{enumitem}
\usepackage{multirow}
\setlist[itemize]{leftmargin=*}

\def\fullfigwidth{1.0\textwidth}
\def\1figwidth{0.55\textwidth}
\def\2figwidth{0.38\textwidth}
\def\3figwidth{0.33\textwidth}
\def\4figwidth{0.24\textwidth}

\newcommand{\bs}[1]{\boldsymbol{#1}}
\newcommand{\mb}[1]{\mathbf{#1}}

\newcommand{\mc}[1]{\mathcal{#1}}

\newcommand{\ctrans}{\mathrm{H}}

\DeclareMathOperator*{\argmin}{arg\,min}
\DeclareMathOperator*{\argmax}{arg\,max}

\def\fig{Fig.~}

\def\secsym{\S}

\def\eg{e.g.}
\def\ie{i.e.}

\def\eqn{Eqn.~}

\def\etc{etc.}
\def\sysname{\textsf{RF-Diffusion}\xspace}

\graphicspath{{./pdf}}

\renewcommand\footnotetextcopyrightpermission[1]{}

\begin{document}
% Title portion
\title{RF-Diffusion: Radio Signal Generation via Time-Frequency Diffusion}

\author{Guoxuan Chi$^1$, {Zheng Yang$^1$\textsuperscript{\Letter}}, Chenshu Wu$^2$, Jingao Xu$^1$, Yuchong Gao$^3$, \\ Yunhao Liu$^1$, Tony Xiao Han$^4$}
\affiliation{%
\institution{$^1${Tsinghua University} \ \ $^2${The University of Hong Kong} \\ $^3${Beijing University of Posts and Telecommunications} \ \ $^4${Huawei Technologies Co., Ltd}}
\country{}
}
\email{{chiguoxuan, hmilyyz, wucs32, xujingao13, gaoyc01, yunhaoliu}@gmail.com, tony.hanxiao@huawei.com}

\renewcommand{\shortauthors}{Guoxuan Chi, Zheng Yang, Chenshu Wu and Jingao Xu, et al.}
\renewcommand{\authors}{Guoxuan Chi, Zheng Yang, Chenshu Wu, Jingao Xu, Yuchong Gao, Yunhao Liu, Tony Xiao Han}

\thanks{\textsuperscript{\Letter} Zheng Yang is the corresponding author.
Our project is available \color{magenta}{\href{https://github.com/mobicom24/RF-Diffusion}{here}}\color{black}{.}}

\input{contents/0-abstract.tex}

\begin{CCSXML}
<ccs2012>
   <concept>
       <concept_id>10003120.10003138</concept_id>
       <concept_desc>Human-centered computing~Ubiquitous and mobile computing</concept_desc>
       <concept_significance>500</concept_significance>
       </concept>
   <concept>
       <concept_id>10003033.10003106.10003113</concept_id>
       <concept_desc>Networks~Mobile networks</concept_desc>
       <concept_significance>500</concept_significance>
       </concept>
 </ccs2012>
\end{CCSXML}

\ccsdesc[500]{Human-centered computing~Ubiquitous and mobile computing}
\ccsdesc[500]{Networks~Mobile networks}

\keywords{RF Signal, Generative Model, Time-Frequency Diffusion, Wireless Sensing, Channel Estimation}

\acmYear{2024}\copyrightyear{2024}
% \setcopyright{acmlicensed}
\acmConference[ACM MobiCom '24]{International Conference On Mobile Computing And Networking}{September 30--October 4, 2024}{Washington D.C., DC, USA}
\acmBooktitle{International Conference On Mobile Computing And Networking (ACM MobiCom '24), September 30--October 4, 2024, Washington D.C., DC, USA}
\acmDOI{10.1145/3636534.3649348}
\acmISBN{979-8-4007-0489-5/24/09}

\maketitle

\input{contents/1-introduction}

\input{contents/2-overview}

\input{contents/3-theory}

\input{contents/4-model}

\input{contents/5-implementation}
\input{contents/6-evaluation}

\input{contents/7-case-study}
\input{contents/8-related-work}
\input{contents/9-discussion}

\input{contents/10-conclusion}

\section{Acknowledgments}
We sincerely thank the MobiSense Group, the anonymous reviewers, and our shepherd for their constructive comments and feedback in improving this work. This paper is supported in part by the NSFC under grant No. 62372265, No.62302254, and No. 62222216, and by the Hong Kong RGC ECS under grant 27204522 and RIF under grant R5060-19.

\appendix
\input{contents/appendix.tex}

\balance
%\end{document}  % This is where a 'short' article might terminate

%ACKNOWLEDGMENTS are optional
% This section is optional; it is a location for you
% to acknowledge grants, funding, editing assistance and
% what have you.  In the present case, for example, the
% authors would like to thank Gerald Murray of ACM for
% his help in codifying this \textit{Author's Guide}
% and the \textbf{.cls} and \textbf{.tex} files that it describes.

%
% The following two commands are all you need in the
% initial runs of your .tex file to
% produce the bibliography for the citations in your paper.
\bibliographystyle{ACM-Reference-Format}
\bibliography{./bib/sigproc}  % sigproc.bib is the name of the Bibliography in this case
% You must have a proper ".bib" file
%  and remember to run:
% latex bibtex latex latex
% to resolve all references
%
% ACM needs 'a single self-contained file'!
%
%APPENDICES are optional

%\balancecolumns

\end{document}

%% file: contents/0-abstract.tex
\begin{abstract}

% \rev{AIGC is revolutionizing the RF domain with its superior RF data generation capability.}
Along with AIGC shines in CV and NLP, its potential in the wireless domain has also emerged in recent years.
% recent years \rev{have also illuminated} its power in the wireless domain.
% The generated RF signal bolsters wireless systems through data augmentation, signal denoising, and channel estimation. 
% \rev{but also streamlines application-layer neural network training by reducing ground truth annotation overheads}.
Yet, existing RF-oriented generative solutions are ill-suited for generating high-quality, time-series RF data due to limited representation capabilities.
%or burdens of model re-training across diverse environments or applications.
In this work, inspired by the stellar achievements of the diffusion model 
%in generating images, videos, and texts 
in CV and NLP, we adapt it to the RF domain and propose \sysname.
To accommodate the unique characteristics of RF signals, we first introduce a novel \textit{Time-Frequency Diffusion} theory to enhance the original diffusion model, enabling it to tap into the information within the time, frequency, and complex-valued domains of RF signals.
%, beyond just the straightforward signal strength.
On this basis, we propose a \textit{Hierarchical Diffusion Transformer} to translate the theory into a practical generative DNN through elaborated design spanning network architecture, functional block, and complex-valued operator, making \sysname a versatile solution to generate diverse, high-quality, and time-series RF data.
% The performance comparison with prevalent generative models including DDPM, DCGAN, and CVAE demonstrates the surperior performance of \sysname in synthesizing Wi-Fi and FMCW signals. We also showcase the versatility of \sysname in boosting Wi-Fi sensing systems and performing channel estimation in 5G networks.
Performance comparison with three prevalent generative models demonstrates the \sysname's superior performance in synthesizing Wi-Fi and FMCW signals. 
We also showcase the versatility of \sysname in boosting Wi-Fi sensing systems and performing channel estimation in 5G networks.
% Our code, data, and pre-trained models are publicly available at \color{magenta}{\url{https://github.com/mobicom24/RF-Diffusion}}.

\end{abstract}

%% file: contents/1-introduction.tex
\section{Introduction}
\label{sec:intro}

Artificial intelligence generated content (AIGC) has catalyzed a revolutionary impact in both industrial and academic frontier, birthing a constellation of cutting-edge products founded on deep neural networks (DNNs).
% Remarkable odysseys include image creation (\eg, Midjourney\tocite, Stable Diffusion\tocite) with GANs or Diffusion models\tocite in the visual domain, and text generation (\eg, ChatGPT\tocite) using Transformer-based large language models\tocite in natural language processing (NLP).
Remarkable odysseys include Stable Diffusion~\cite{rombach2022high},  Midjourney~\cite{midjourney}, DALL-E~\cite{ramesh2022hierarchical} for image creation, and ChatGPT~\cite{openai2023gpt4} for text generation.
% The ascent of AIGC is typically remarked as ``comparable to an Industrial Revolution''~\cite{su2023research}.

Nowadays, AIGC is gradually knocking on the door of the radio-frequency (RF) domain.
Current practice offers initial proof of its potential to boost wireless systems in terms of data augmentation~\cite{rizk2019effectiveness}, signal denoising~\cite{bando2020adaptive} and time-series prediction~\cite{hamdan2020variational}. 
In downstream tasks like device localization~\cite{zhao2023nerf}, human motion sensing~\cite{yang2022rethinking}, and channel estimation~\cite{liu2021fire}, such progress not only enhances system performance but also cuts down the cumbersome ground truth annotation costs for application-layer DNN training.
% costs tied to labeling data for neural network training.

Existing RF data generation models can be broadly divided into two main categories:

% \noindent \textbf{$\bullet$ Physical simulation based generation model.}\rev{physical-model-driven generative model, physics-based model.}
\noindent \textbf{$\bullet$ Environment modeling based generative model.}
% This approach exploits extra LiDAR point clouds or camera-captured videos to create a detailed 3D model of the environment and target. 
% It then applies physical-level propagation models of electromagnetic (EM) signal (\eg, ray tracing model\tocite, radar communication principle\tocite) to simulate how signals interact with surroundings and obstacles, which helps eventually predict the signals a receiver would sample.
This approach exploits LiDAR point clouds or video footage to craft a detailed 3D model of the environment. 
It then employs physical models, like ray tracing~\cite{mckown1991ray}, to simulate how RF signals interact with surroundings, which eventually aids in forecasting the signals a receiver might capture.
However, one notable limitation is the method's insufficient consideration of how the materials and properties of targets can affect RF signal propagation.
Additionally, obtaining a 3D model with accuracy compatible with RF signal wavelengths (\eg, 1-10 mm) remains a challenge and will significantly raise system expenses.
While the recent study uses the neural radiance field for implicit modeling of RF-complex environments to estimate signal propagation~\cite{zhao2023nerf}, it requires a stationary receiver (Rx), which complicates generating essential time-series data for wireless communication systems or tasks like human motion recognition.

% \noindent \textbf{$\bullet$ Data distribution based generation model.} \rev{Probabilistic Generative Model, data-driven generative model.}
\noindent \textbf{$\bullet$ Data-driven probabilistic generative model.}
Current innovations leverage models like generative adversarial network (GAN) and variational autoencoder (VAE) to augment RF datasets~\cite{ha2020food}. 
Essentially, these models learn the distribution within the training data and then generate new RF data that follow this distribution.
However, these models mainly focus on expanding feature-level distributions and struggle to precisely generate raw RF signals due to their constrained representation capabilities~\cite{yang2022rethinking}.
Additionally, 
% they require adjustments in loss functions and architectures for specific tasks, limiting their versatility\tocite. \rev{
most of them are designed for specific tasks with dedicated loss functions and DNN architectures, limiting their versatility.
On the other hand, GAN's training is notoriously fickle due to the tug-of-war between the generator and discriminator~\cite{xia2022gan}.
% , \rev{requiring a large training dataset to function effectively, which contradicts the goal of efficient data generation}\tocite.

% The most recent work, NeRF$^{2}$, adapts the neural radiance field (NeRF) from optics to the RF domain, enabling accurate estimation of RF signal's propagation\tocite.
% \rev{However, NeRF$^{2}$ requires a stationary receiver (Rx), which complicates generating time-series data essential for tasks like behavior and motion recognition \rev{and wireless communication systems}}.
% Moreover, it can only generate data in a previously trained environment, emphasizing the data's domain-dependent property and hindering models' generalizability.
% transformation capability \rev{generalizability}.

% Motivating Example
% 1. 我们的频谱精细（特征精细），因为时频+diffusion
% 2. DDPM频谱不精细，因为没有时频；
% 3. VAE最不精细，因为没有diffusion，也没有时频，没有insight；
% 4. GAN看起来还行，但是是因为模式坍塌，生成的样本基本都是一个样子，不好不坏

% \rule[0pt]{0mm}{1cm}\includegraphics[scale=0.1]{pdf/00-wifi-gt.png}

% Please add the following required packages to your document preamble:
% \usepackage{multirow}
\begin{table}[]
\setlength\abovecaptionskip{0pt}
\caption{Illustrative examples.}
\begin{threeparttable}
\begin{tabular}{m{1.2cm}<{\centering} m{1.8cm}<{\centering} m{1.8cm}<{\centering} m{0.0cm}<{\centering} m{0.9cm}<{\centering} m{0.8cm}<{\centering}}
\hline
\multirow{2}{*}{Methods} & \multicolumn{2}{c}{Examples} & \multirow{2}{*}{} & \multicolumn{2}{c}{SSIM} \\ \cline{2-3} \cline{5-6} 
                         & Wi-Fi         & FMCW         &                   & Wi-Fi       & FMCW       \\ \hline
Ground
Truth              & \rule[0pt]{0mm}{1.5cm}\includegraphics[scale=0.16]{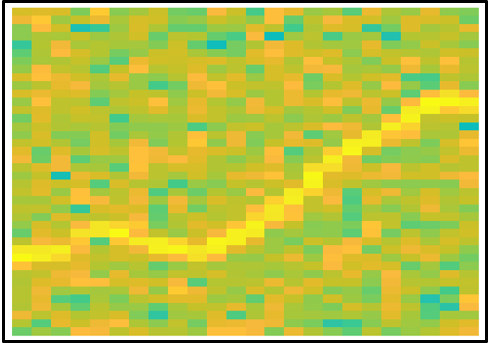}   &  \rule[0pt]{0mm}{1.4cm}\includegraphics[scale=0.16]{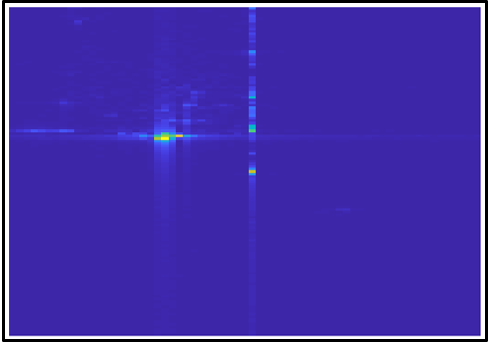}         &  &     N/A        &     N/A    \\
\textbf{Ours}    & \rule[0pt]{0mm}{1.1cm}\includegraphics[scale=0.16]{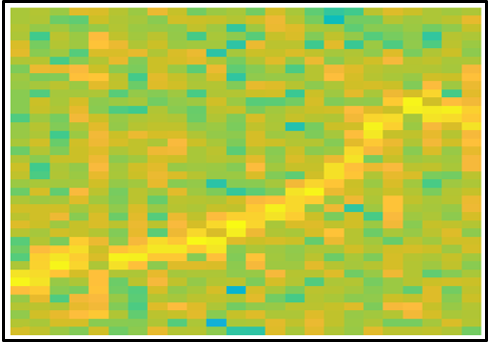}   &  \rule[0pt]{0mm}{1.1cm}\includegraphics[scale=0.16]{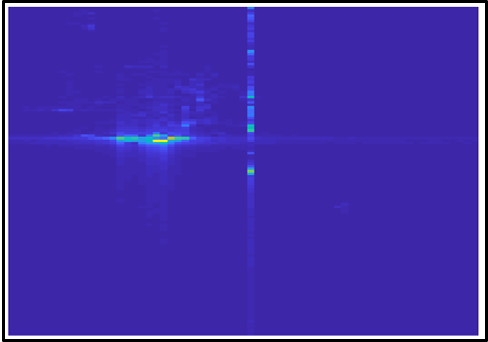}         &  &     \textbf{0.81}        &     \textbf{0.75}       \\
DDPM
\cite{ho2020denoising}                   & \rule[0pt]{0mm}{1.1cm}\includegraphics[scale=0.16]{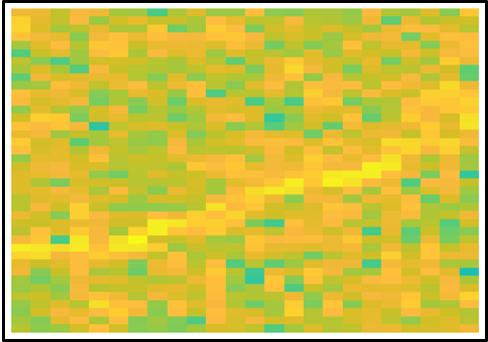}   &  \rule[0pt]{0mm}{1.1cm}\includegraphics[scale=0.16]{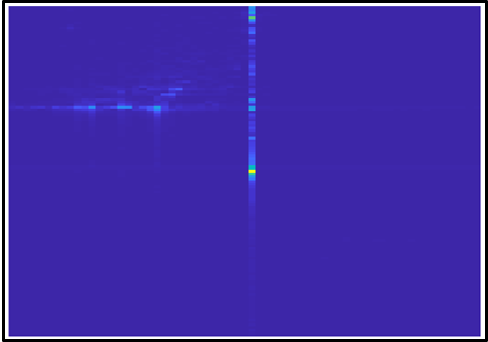}         &  &     0.65        &     0.58       \\
DCGAN
\cite{radford2015unsupervised}                 & \rule[0pt]{0mm}{1.1cm}\includegraphics[scale=0.16]{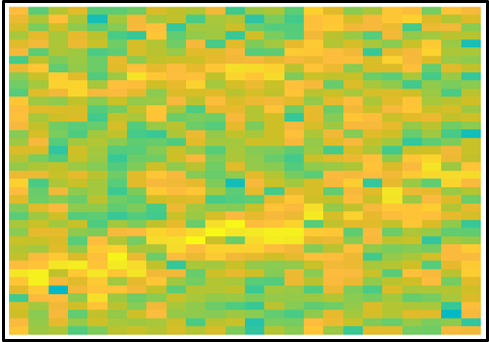}   &  \rule[0pt]{0mm}{1.1cm}\includegraphics[scale=0.16]{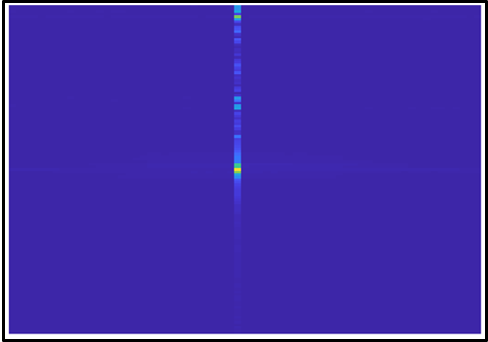}         &  &     0.68        &     0.61       \\
CVAE
\cite{sohn2015learning}                    & \rule[0pt]{0mm}{1.1cm}\includegraphics[scale=0.16]{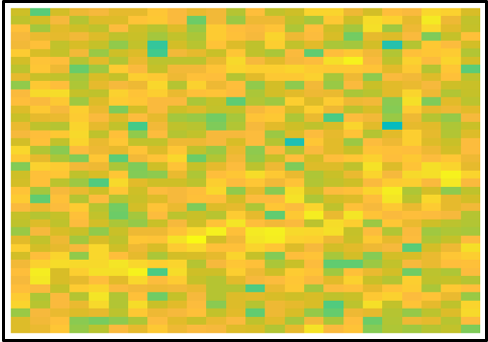}   &  \rule[0pt]{0mm}{1.1cm}\includegraphics[scale=0.16]{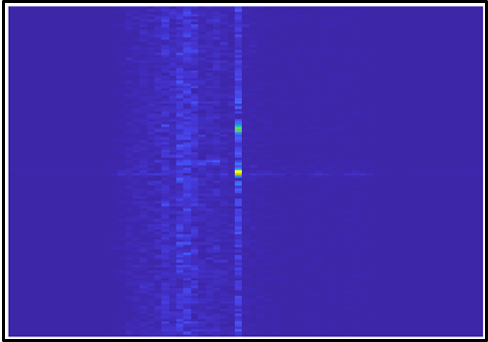}         &  &     0.47        &     0.4         \\ \hline
\end{tabular}
% \begin{tablenotes}
% \item[-]\small{A comparison among \sysname, DDPM (original Diffusion-based), DCGAN (GAN-based), and CVAE (VAE-based). \sysname achieves the best raw signal generation similarity.}
% \end{tablenotes}
\end{threeparttable}
\label{tab:examples}
\end{table}

\noindent \textbf{Remark.}
Albeit inspiring, there still lacks a versatile generative model for generating accurate and time-series raw RF signals suitable for diverse downstream applications.

% AIGC model in the RF domain.
Recently, \textit{Diffusion Model} has emerged as a luminous star in the visual AIGC cosmos, underpinning a variety of innovative DNNs for a range of prominent image/video applications such as Stable Diffusion, Midjourny, and DALL-E.
Compared to the aforementioned generative models, its unique iterative process of noise addition (\ie, noising) and removal (\ie, denoising) allows for precise capture of intricate raw data distributions~\cite{yang2022diffusion}.
Moreover, its training is straightforward and avoids typical problems like mode collapse or convergence troubles, since it doesn't juggle competing parts or require delicate fine-tuning~\cite{dhariwal2021diffusion}.
% without the need for balancing competing components or fine-tuning intricate encodings, the training remains more predictable and less prone to common pitfalls such as mode collapse or convergence issues.
% bedrock for a myriad of avant-garde applications.
% has achieved remarkable success in the visual AIGC domain, becoming the foundational network for numerous leading-edge applications.

These compelling advantages inspire us to embrace the diffusion model for synthesizing RF data.
However, transferring existing diffusion models~\cite{ho2020denoising} to the RF domain faces significant challenges arising from RF signal's unique characteristics beyond images, as summarized below.

% \begin{itemize}
% \item 
\noindent $(i)$ \textit{Time series}. RF signals capture dynamic details like target movement and environment/channel changes over time, unlike static snapshots.
Diffusion models designed for single-image generation struggle to synthesize RF signal sequences.

\noindent $(ii)$ \textit{Frequency domain}. Essential RF details (\eg, Doppler shift, chirp) are embedded in the frequency domain. 
While recent video diffusion models can create time series, they mainly focus on the spatial domain (\eg, pixel-wise brightness), discarding the rich information in the frequency domain.

\noindent $(iii)$ \textit{Number field}. RF data is complex-valued with both amplitude and phase readings.
While existing diffusion models only focus on amplitude (\eg, light strength), the phase data can't be ignored due to its crucial role in wireless systems.
% \end{itemize}

% \vspace{-2mm}
\noindent In summary, while diffusion models hold great promise, there is a need to upgrade current models to suit the unique traits of RF signals and tap into the underlying information in the time series, frequency, and complex-valued domains.
% to produce high-quality, diverse, and time-series-information-rich RF data for a broad range of wireless applications 
% is non-trivial and 
% faces significant challenges - 
% the inherent traits of RF signals, such as their unique characteristics in \textit{time} series, \textit{frequency} domain, and \textit{number} field, present substantial hurdles that differ from those in image synthesising, as summarized below.

% Firstly, RF signals capture dynamic details like target movement and environment/channel changes over time, unlike static snapshots.
% Diffusion models designed for single-image generation struggle to synthesis continuous RF signal sequences.
% Secondly, essential details (\eg, Doppler shift, chirp) are typically embedded in the frequency domain of RF signals. 
% While recent video and audio diffusion models can create time series, they mainly focus on the spatial domain (\ie, pixel brightness or sound intensity), missing out on the rich information in the frequency domain.
% Lastly, RF data is complex-valued with both amplitude and phase readings.
% While existing diffusion models only focus on amplitude (\ie, light strength), the phase, especially in \rev{short-wavelength} RF signals, is vital and can't be ignored due to its crucial role in wireless systems\tocite.
% Different from existing diffusion models where only the amplitude (\ie, signal strength) is considered, the phase in cm- or mm-wavelength RF signals cannot be disregarded for its crucial role in wireless systems\tocite.

% \com{I'm a little confused by the current description of challenges.}

\vspace{1mm}
\noindent \textbf{Our Work.}
% To overcome the above challenges, 
We propose \sysname, the first versatile generative model for \textbf{RF} signals based on \textbf{Diffusion} model.
To overcome the above challenges, we expand existing denoising-based diffusion model to the time-frequency domain by revisiting its theoretical foundation, overall DNN architecture, and detailed operator design, enabling \sysname to generate diverse, high-quality, and time-series RF data.
% expand existing 
% \sysname is designed to generate diverse, high-quality, and time-series RF data for a wide spectrum of upper-layer wireless applications.
% \com{use 1-2 sentences to introduce the input and output of \sysname. \eg, NeRF2 can tell how/what signal is received at any position when it knows the position of a transmitter}.
% To overcome the above challenges, \sysname excels in two aspects:

% \noindent \textbf{$\bullet$ On the RF-oriented Diffusion Theory front.}
\noindent \textbf{$\bullet$ Time-Frequency Diffusion Theory.}
We first propose the \textit{time-frequency diffusion} (TFD) theory as a novel paradigm to guide diffusion models in extracting and leveraging characteristics of RF signals across both temporal and frequency domains.
% update the original diffusion model to fit the time-frequency characteristic of RF signals.
% , emphasizing details in all the spatial, temporal, and frequency domain.
% In TFD, 
Specifically, we demonstrate a diffusion model could effectively destruct and restore high-quality RF signals by alternating between adding noise in the time domain and blurring in the frequency domain (\S\ref{sec:theory}).

% \noindent \textbf{$\bullet$ On the Diffusion Model Design front.}
\noindent \textbf{$\bullet$ Hierarchical Diffusion Transformer Design.}
We further re-design the DNNs of existing denoising-based diffusion model to be compatible with TFD.
The derived DNN, designated as the \textit{hierarchical diffusion transformer} (HDT), from a top-down perspective, 
% and propose \textit{hierarchical diffusion transformer} (HDT).
% % We \rev{translate} the above TFD theory into a practical generative DNN, namely \textit{hierarchical diffusion transformer} (HDT).
% In general, HDT 
incorporates $(i)$ a hierarchical architecture to fully uncover time-frequency details by decoupling spatio-temporal dimensions of RF data;
$(ii)$ attention-based diffusion blocks leveraging enhanced Transformers to extract RF features;
and $(iii)$ a complex-valued design to encode both signal strength and phase information.
The three key designs work hand-in-hand to enable \sysname to generate high-quality RF data (\S\ref{sec:model}).

We implement \sysname and conduct extensive experiments that include synthesis of both Wi-Fi and FMCW signals. To provide a clear understanding of its performance, an intuitive comparison of the time-frequency spectrograms generated by \sysname and those from related works is presented in Table \ref{tab:examples}.
Evaluation results demonstrate that \sysname generates RF signals with high fidelity, achieving an average structural similarity of $81\%$ relative to the ground truth. This performance surpasses prevalent generative models such as DDPM, DCGAN, and CVAE by over $18.6\%$.
We also demonstrate the performance of \sysname in two case studies: augmented Wi-Fi gesture recognition and 5G FDD channel estimation. 
By employing \sysname as a data augmentor, existing wireless gesture recognition systems experience a significant accuracy improvement ranging from $4\%$ to $11\%$. When applied to the channel estimation task, \sysname showcases a substantial $5.97$ dB improvement in SNR compared to state-of-the-arts.

In summary, this paper makes the following contributions.

\noindent (1) We propose \sysname, the first generative diffusion model tailored for RF signal.
\sysname is versatile and can be leveraged in a wide spectrum of fundamental wireless tasks such as RF data augmentation, channel estimation, and signal denoising, propelling AIGC to shine in the RF domain.

% \noindent (2) We present \textit{Time-Frequency Diffusion}, a comprehensive diffusion modeling approach for the RF domain, integrating theory foundations, network architecture, and functional design.
% As a result, \sysname excels in generating high-quality, diverse, and time-series RF data.

\noindent (2) 
% We introduce the \textit{Time-Frequency Diffusion} theory, a significant upgrade of the current denoising-based diffusion ones.
% TFD, coupled with its compatible \textit{Hierarchical Diffusion Transformer} DNN, empowers a diffusion model to precisely handle time-series sampling while equally emphasizing the data's spectral details.
We present the \textit{Time-Frequency Diffusion} theory, an advanced evolution beyond traditional denoising-based diffusion methods. The integration of TFD with its bespoke \textit{Hierarchical Diffusion Transformer} (HDT) enables enhanced precision in time-series sampling and a balanced focus on spectral details of the data.

\noindent (3) We fully implement \sysname. Extensive evaluation results from case studies show \sysname's efficacy. 

\noindent \textbf{Community Contribution.}
% First, we introduce the \textit{Time-Frequency Diffusion} theory, a significant \rev{enhancement} of the current denoising-based diffusion ones.
% TFD enables a diffusion model to precisely handle time-series sampling while equally emphasizing the data's spectral details.
% Its potential extends beyond the wireless community, offering value to video, audio processing, and other time-series-dependent domains \rev{modalities}.
% Second, 
\sysname's code and pre-trained model are publicly available.
Our solution, in part or in all, could provide a collection of tools for both industry and academia to push forward AIGC in RF domain.
Moreover, its ability to handle time-series sampling while highlighting the spectral nuances of the data has potential benefits beyond the wireless community, offering value to video, audio processing, and other time-series-dependent modalities.

% Its capability to handle time-series sampling while equally emphasizing the data's spectral details also hold potential beyond the wireless community, offering value to video, audio processing, and other time-series-dependent modalities.

%% file: contents/2-overview.tex
\vspace{-10pt}
\section{Overview}
\label{sec:overview}

\begin{figure}[t]
    \centering
    \setlength\abovecaptionskip{5pt}
    \includegraphics[]{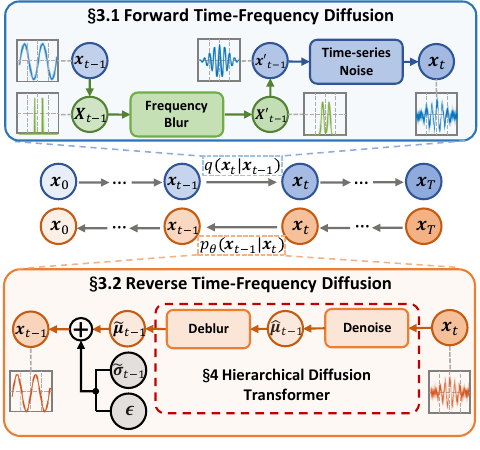}
    \caption{\sysname overview.}
    \Description{}
    \label{fig:overview}
\end{figure}

% \subsection{Problem Statement}
% We propose \sysname, a probabilistic generative model for RF data based on the diffusion model as illustrated in \fig\ref{fig:overview}.
% From a top-level perspective, \sysname shares a similar concept with existing denoising-based diffusion models -- both involve a forward process (\ie, adding noise to data) and a reverse process (\ie, generating data from noise).
% Nonetheless, there are two highlights in \sysname:

We propose \sysname, a pioneering probabilistic generative model for RF data that leverages the diffusion model framework, as detailed in \fig\ref{fig:overview}. At its core, \sysname aligns with the principle of denoising-based diffusion models by employing a dual-process approach: a forward process of integrating noise into the data, and a reverse process of generating data from noise. However, \sysname distinguishes itself through two innovative features:

\noindent $(i)$ \sysname incorporates the proposed Time-Frequency Diffusion (\S\ref{sec:theory}) theory to direct each stage of state transition in both forward (\ie, $q(\boldsymbol{x}_t \vert \boldsymbol{x}_{t-1})$) and reverse (\ie, $p_{\theta}(\boldsymbol{x}_{t-1} \vert \boldsymbol{x}_{t})$) processes, enabling \sysname to harness the RF signal information across both the time and frequency domain.

\noindent $(ii)$ 
% \sysname introduces Hierarchical Diffusion Transformer (\S\ref{sec:model}), restructuring the DNN of the reverse generation process to align with the TFD theory and the traits of RF signals.
\sysname introduces the Hierarchical Diffusion Transformer (\S\ref{sec:model}), which is a restructured DNN model for the reverse generation process, to align with the Time-Frequency Diffusion theory and the characteristics of RF signals.

% From a top-level perspective, we propose a probabilistic generative model called \sysname.
% As illustrated in \fig\ref{fig:overview}, \sysname employs a time-frequency diffusion (\secsym\ref{sec:theory}) process to convert any RF data into tractable noise. Subsequently, a hierarchical diffusion transformer model (\secsym\ref{sec:model}) is leveraged to reverse this process, thereby generating high-fidelity RF data from noise.
% sysname shares the same basic principle with traditional diffusion models~\cite{ho2020denoising, nichol2021improved}. Specifically, in \sysname, 

% \sysname focuses on the conditional generation task, which means it is capable of learning the conditional distribution $q(\boldsymbol{x} \vert \boldsymbol{c})$ of wireless data, and thus generating a specific type of wireless signal given a condition $\boldsymbol{c}$.

% \subsection{\sysname Design}

% 具体来讲，时频diffusion同时关注重建无线信号的时域幅度准确性以及频域序列连续性，并在前向退化过程中联合执行两类操作：在时域添加高斯噪声、在频域进行频谱模糊。随着diffusion step t的增加，原始无线信号的数据分布将被逐渐drown out，并最终使信号退化为一个非各向同性的噪声分布。根据时频diffusion理论，被退化到任意程度的信号xt均能够通过一个逆向重建过程恢复至原始信号x0。同时，由于在重建过程中同时考虑了时频域的准确性，重建的信号是时频高保真的。
% 被精心设计以用于拟合该逆向过程的层级式transformer架构能够高效地解耦信号中的噪声与频谱模糊，并迭代式地清除。
As for the specific data flow, \sysname gradually introduces Gaussian noise in the time domain and blurs the spectrum in the frequency domain at each stage in the forward direction.
As the diffusion step $t$ advances, the original RF signal $\boldsymbol{x}_0$ diminishes, eventually degrading into noise.
In TFD theory, we demonstrate any destructed signal $\boldsymbol{x}_t$ can be restored to its original form $\boldsymbol{x}_0$ using a parameterized reverse process. 
Guided by the destruction process alternating in time-frequency domain, the reverse restoration process emphasizes both time-domain amplitude accuracy and frequency-domain continuity to achieve time-frequency high-fidelity signal generation.

% Specifically, the time-frequency diffusion
% emphasizes both the amplitude accuracy and the sequence continuity of RF signal, and thus jointly performs two types of operations in the forward destruction process: introducing Gaussian noise in the time domain and blurring the spectrum in the frequency domain.  As the diffusion step $t$ advances, the original RF signal $\boldsymbol{x}_0$ diminishes, eventually degrading into noise. According to 
% our time-frequency diffusion theory, any destructed signal $\boldsymbol{x}_t$ can be restored to its original form $\boldsymbol{x}_0$ using a parameterized reverse process. Guided by the destruction process alternating in time-frequency domain, the reverse restoration process emphasizes both time-domain amplitude accuracy and frequency-domain continuity to achieve time-frequency high-fidelity signal generation.

In the reverse direction, HDT are served
% We propose the hierarchical diffusion transformer (HDT) 
as the parameterized model for learning the restoration process. 
It decouples the Gaussian noise and the spectral blur, effectively addressing them in the spatial denoise and time-frequency deblur stages, respectively. 
During its training, HDT takes destructed signal $\boldsymbol{x}_{t}$
as the model input, and uses the signal of previous diffusion step $\boldsymbol{x}_{t-1}$ to supervise the output. Once trained, \sysname is capable of iteratively transforming fully degraded noise back into a specific type of signal. 

% The hierarchical diffusion transformer is elaborately designed as the parameterized model to learn the restoration process. By decoupling the Gaussian noise and spectral blur, HDT effectively eliminates them in the spatial denoise and time-frequency deblur stage respectively.
% % Our proposed hierarchical diffusion transformer is elaborately designed to effectively learn the restoration process by decoupling the Gaussian noise and spectral blur, and then eliminates them in the spatial denoise and time-frequency deblur stage respectively.
% During the training process, HDT takes destructed signal $\boldsymbol{x}_{t}$
% %, its diffusion step $t$, and the associated condition label $\boldsymbol{c}$ 
% as the model inputs, and uses the signal of previous diffusion step $\boldsymbol{x}_{t-1}$ to supervise the output. Once trained, \sysname is capable of transforming a fully-degraded noise into a specific type of signal. 
%$\boldsymbol{c}$, achieving the task of conditional generation of wireless signal.

%% file: contents/3-theory.tex
\section{Time-Frequency Diffusion}
\label{sec:theory}

% \begin{figure}[t]
%     \centering
%     \setlength\abovecaptionskip{5pt}
%     \includegraphics[]{pdf/03-1-time-frequency.pdf}
%     \caption{Diffusion process in the time and frequency domain.}
%     \Description{}
%     \label{fig:diffusion}
% \end{figure}

In this section, we introduce the proposed Time-Frequency Diffusion (TFD) process. 
% As an iterative generative model, TFD shares a fundamental principle with the classic denoising diffusion model~\cite{ho2020denoising, nichol2021improved}. 
% As depicted in \fig\ref{fig:overview}, during the forward destruction process, information in the original signals are gradually eliminated. Subsequently, by reversing the destruction process, new signals can be generated.
Unlike prevailing denoising diffusion models, the time-frequency diffusion process comprehensively addresses two potential distortions in wireless signal data: 1) amplitude distortion due to additive Gaussian noise; 2) spectral aliasing resulting from insufficient time resolution. Therefore, the learned reverse process focuses not only on precisely reconstructing the amplitude of individual samples but also on preserving spectral resolution in time-series signals.
In what follows, we first introduce the forward destruction process (\secsym \ref{subsec:forward}) which jointly eliminates the original data distribution in the time and the frequency domain. On this basis, we describe how to reverse the process (\secsym \ref{subsec:reverse}) and fit it through a parameterized model, which is the basis of our conditional generation (\secsym \ref{subsec:cond}) task.
% which is capable of generating high-fidelity signal from the noise.
% \vspace{-5pt}
\subsection{Forward Destruction Process}
\label{subsec:forward}
% TODO 可能还不够清楚
The time-frequency diffusion model is proposed for the RF signal, which can be treated as the complex-valued time-series data. Therefore, we take the signal as a two-dimensional complex tensor $\bs{x} \in \mathbb{C}^{M \times N}$, where $M$ represents the spatial dimension of each sample, while $N$ represents the temporal dimension of the times series.

Given a signal that follows a specific distribution $\bs{x}_0 \sim q(\bs{x}_0)$, the forward destruction process yields a progression of random variables $\bs{x}_1, \bs{x}_2, \dots, \bs{x}_T$. Each diffusion step in this process disrupts the original distribution from both the time and frequency domains. Specifically, the forward diffusion process from step $t-1$ to $t$ is described as follows:
\begin{itemize}
    \item \textbf{Frequency Blur.} To dissipate the spectral details of the original signal, the Fourier transform $\mathfrak{F}(\cdot)$ is first performed to the temporal dimension. Subsequently, with the predefined Gaussian convolution kernel $\mb{G}_{t}$, a cyclic convolution $*$ operation is performed on the spectrum, resulting in a blurred spectrum $\mathbf{G}_{t} * \mathfrak{F}(\boldsymbol{x}_{t-1})$.
    \item \textbf{Time-series Noise.} To drown out the amplitude details of the signal, complex standard Gaussian noise $\bs{\epsilon} \sim \mc{CN}(0, \mb{I})$ is introduced, and a weighted summation is performed with a predefined parameter $\sqrt{\alpha_t}$, where $\alpha_t \in (0, 1)$.
\end{itemize}
By combining the above two steps, we get:
\begin{equation}
\label{eqn:forward}
\bs{x}_{t}=\sqrt{\alpha_t}\mathfrak{F}^{-1}(\mathbf{G}_{t} * \mathfrak{F}(\boldsymbol{x}_{t-1})) + \sqrt{1-\alpha_t} \boldsymbol{\epsilon},
\end{equation}
where $\mathfrak{F}^{-1}(\cdot)$ indicates the inverse Fourier transform. 

% Based on the chain rule, the joint distribution of $\boldsymbol{x}_1, \boldsymbol{x}_2, \dots, \boldsymbol{x}_T$ conditioned on the original information $\boldsymbol{x}_0$, can be written as:
% \begin{equation}
%   q(\boldsymbol{x}_1, \boldsymbol{x}_2, \dots, \boldsymbol{x}_T \vert \boldsymbol{x}_0) = \prod_{t=1}^{T} q(\bs{x}_t \vert \bs{x}_{t-1}),
% \end{equation}

% The aforementioned process can be regarded as a Markov chain, with the transition probability from step $t-1$ to $t$ is formulated as follows:
% \begin{equation}
%     q(\boldsymbol{x}_t \vert \boldsymbol{x}_{t-1}) = \mathcal{CN}(\boldsymbol{x}_t; \sqrt{\alpha_t} \mathfrak{F}^{-1}(\mathbf{G}_{t} * \mathfrak{F}(\boldsymbol{x}_{t-1})), (1 - \alpha_t)\mathbf{I}).
% \end{equation}

To ensure the practical feasibility of the time-frequency diffusion process, it is essential that the transition from $\bs{x}_0$ to $\bs{x}_t$ for any given step $t\in [1, T]$ can be executed with an acceptable time complexity, instead of involving an iteration of $t$ steps. To simplify this process, certain advantageous characteristics of the Fourier transform and the Gaussian function are leveraged.
% Some favorable properties of the Fourier transform and Gaussian function are leveraged to simplify the aforementioned process. 
Based on the convolution theorem~\cite{conv}, we have $\mathfrak{F}^{-1}(\mathbf{G}_{t} * \mathfrak{F}(\boldsymbol{x}_{t-1})) = \mathfrak{F}^{-1}(\mathbf{G}_{t}) \boldsymbol{x}_{t-1}$~\footnote{Vector multiplications in this paper default to element-wise products.}. Therefore, the operation in \eqn\ref{eqn:forward} can be expressed as:
\begin{equation}
\begin{aligned}
\label{eqn:forward-t-1-t}
\boldsymbol{x}_t & = \sqrt{\alpha_t}\boldsymbol{g}_{t}  \boldsymbol{x}_{t-1} + \sqrt{1-\alpha_t} \boldsymbol{\epsilon}, \\
\end{aligned}
\end{equation}
where $\boldsymbol{g}_t = \mathfrak{F}^{-1}(\boldsymbol{G}_t)$ is still a Gaussian kernel, which means the convolution of the signal with the Gaussian kernel in the frequency domain can be equivalently transformed into the multiplication of the signal with another Gaussian kernel in the time domain. For ease of notion, let \( \boldsymbol{\gamma}_t = \sqrt{\alpha_t}\boldsymbol{g}_{t} \), and \( \sigma_{t} = \sqrt{1 - \alpha_t} \), indicating the weight of the signal $\bs{x}_{t-1}$ and the standard deviation of the added noise at step \( t \).

% \begin{equation}
% \begin{aligned}
% \boldsymbol{x}_t & = \sqrt{\alpha_t}\boldsymbol{g}_{t} \circ \boldsymbol{x}_{t-1} + \sqrt{1-\alpha_t} \boldsymbol{\epsilon} \\
% & = \boldsymbol{\gamma}_{t} \circ \boldsymbol{x}_{t-1} + \sigma_{t}\boldsymbol{\epsilon}
% \end{aligned}
% \end{equation}

% where \( \boldsymbol{\gamma}_t = \sqrt{\alpha_t}\boldsymbol{g}_{t} \) and \( \sigma_{t} = \sqrt{1 - \alpha_t} \).

% Specifically, considering that the Fourier transform of a Gaussian function is still a Gaussian function, and the product of Gaussian functions remains a Gaussian function, \( \boldsymbol{g}_{t} \) is also a Gaussian function. 
Since the forward process is a Markov chain, by recursively applying \eqn\ref{eqn:forward-t-1-t} and incorporating with the reparametrization trick~\cite{kingma2013auto}, the relationship between the original signal $\bs{x}_0$ and the degraded signal $\bs{x}_t$ can be obtained: 
\begin{equation}
\begin{aligned}
\label{eqn:forward-0-t}
\boldsymbol{x}_t & = \bar{ \boldsymbol{\gamma}}_{t} \boldsymbol{x}_{0}  + \sum_{s = 1}^{t} {(\sqrt{1 - \alpha_s} \frac{\bar{\boldsymbol{\gamma}}_{t}}{\bar{\boldsymbol{\gamma}}_{s}}) \boldsymbol{\epsilon}} =  \bar{\boldsymbol{\gamma}}_{t}  \boldsymbol{x}_{0} + \bar{\boldsymbol{\sigma}}_{t} \boldsymbol{\epsilon},
\end{aligned}
\end{equation}
where \( \bar{\boldsymbol{\gamma}}_{t} = \prod_{s = 1}^{t} \bs{\gamma}_{s} =  \boldsymbol{\gamma}_t \cdots \boldsymbol{\gamma}_{1} \).
% Where \(\bar{\boldsymbol{\gamma}}_{t}\) represents the proportion of the original signal \(\bs{x}_0\) in \(\bs{x}_t\), and \(bar{\boldsymbol{\sigma}}_{t}\) represents the standard deviation of the Gaussian noise in \(\bs{x}_t\).
% Therefore, let's define 
% \[
% \bar{\boldsymbol{\gamma}}_{t} = \boldsymbol{\gamma}_t \circ \cdots \circ \boldsymbol{\gamma}_{1}
% \]
As \( \alpha_t \) and \( \boldsymbol{g}_t \) are predefined hyperparameters corresponding to the noise and blur scheduling strategy, any \( \bar{\boldsymbol{\gamma}}_{t} \) and \( \bar{\boldsymbol{\sigma}}_{t} \) are constant coefficients, representing the weight of the original signal and the standard deviation of the added noise. Thus, the forward destruction process to any step \( t \) can be quickly completed without iteration. 
Stated in probabilistic terms, essentially \( \boldsymbol{x}_t \) follows an non-isotropic Gaussian distribution conditioned on \( \boldsymbol{x}_0 \):
\begin{equation}
\label{eqn:forward-prob}
 q(\boldsymbol{x}_t \vert \boldsymbol{x}_0) = \mathcal{CN}(\boldsymbol{x}_0; \bar{\boldsymbol{\mu}}_{t}, \bar{\boldsymbol{\sigma}}_{t}^{2} \mathbf{I}),  
\end{equation}
where \( \bar{\boldsymbol{\mu}}_{t} = \bar{\boldsymbol{\gamma}}_{t} \boldsymbol{x}_{0} \) and \( \bar{\boldsymbol{\sigma}}_{t} = \sum_{s = 1}^{t} {(\sqrt{1 - \alpha_s} \frac{\bar{\boldsymbol{\gamma}}_{t}}{\bar{\boldsymbol{\gamma}}_{s}}}) \). Specifically, the vector \( \bar{\boldsymbol{\gamma}}_{t} \) consists of distinct weighting coefficients, each applied multiplicatively across the temporal dimension of the original signal to perform weighting adjustments.

It is proven in Appendix~\ref{sec:proof-1} that as the diffusion step \( t \) increases, the original signal is gradually eliminated, and \( \bs{x}_t \) eventually converges to a closed-form noise distribution:
\begin{equation}
\label{eqn:forward-lim}
\lim_{T \to \infty} \bs{x}_T = \lim_{T \to \infty} \sum_{t = 1}^{T} {(\sqrt{1 - \alpha_t}\frac{\bar{\boldsymbol{\gamma}}_{T}}{\bar{\boldsymbol{\gamma}}_{t}}}) \boldsymbol{\epsilon} = \lim_{T \to \infty} \bar{\boldsymbol{\sigma}}_{T}  \boldsymbol{\epsilon},
\end{equation}
where $\bar{\boldsymbol{\sigma}}_{T} =  \sum_{t = 1}^{T} {(\sqrt{1 - \alpha_t}\frac{\bar{\boldsymbol{\gamma}}_{T}}{\bar{\boldsymbol{\gamma}}_{t}}}) \boldsymbol{\epsilon}$ is determined by predefined noise scheduling strategy in practical implementation.

\subsection{Reverse Restoration Process}
\label{subsec:reverse}

The restoration process is the reversal of the destruction, which gradually eliminates the noise and restores the original data distribution.

To learn a parameterized distribution $p_{\theta}(\bs{x}_0)$ which approximates the original distribution $q(\bs{x}_0)$, an effective approach is to minimize their Kullback-Leibler (KL) divergence:

\begin{equation}
\begin{aligned}
\label{eqn:opt-kl}
\theta&=\argmin_{\theta}D_{\mathrm{KL}}(q(\boldsymbol{x}_0)\Vert p_{\theta}(\boldsymbol{x}_0))\\
&=\argmin_{\theta}(\mathbb{E}_{q(\bs{x}_0)}[-\log p_{\theta}(\boldsymbol{x}_0)]+\mathbb{E}_{q(\bs{x}_0)}[\log q(\boldsymbol{x}_0)])\\
&=\argmax_{\theta}\mathbb{E}_{q(\bs{x}_0)}[\log p_{\theta}(\boldsymbol{x}_0)].
% &\approx\argmin_{\theta}D_{\mathrm{KL}}(q(\boldsymbol{x}_{t-1}\vert\boldsymbol{x}_{t}, \boldsymbol{x}_{0})\Vert{p_{\theta}(\boldsymbol{x}_{t-1} \vert \boldsymbol{x}_{t})}).
\end{aligned}
\end{equation}
Unfortunately, $q(\bs{x}_0)$ is intractable to calculate in general~\cite{sohl2015deep, kong2020diffwave}, therefore \(\mathbb{E}_{q(\bs{x}_0)}[\log p_{\theta}(\boldsymbol{x}_0)] \) cannot be expressed explicitly. Building on the concepts of prior works~\cite{song2019generative, ho2020denoising}, we approximate the distribution by maximizing the variational lower bound. As established in~\cite{ho2020denoising}, the optimization problem in \eqn\ref{eqn:opt-kl} can be approximated as:
\begin{equation}
\label{eqn:opt-vlb}
    \theta = \argmin_{\theta}D_{\mathrm{KL}}(q(\boldsymbol{x}_{t-1}\vert\boldsymbol{x}_{t}, \boldsymbol{x}_{0})\Vert{p_{\theta}(\boldsymbol{x}_{t-1} \vert \boldsymbol{x}_{t})}),
\end{equation}
where $q(\boldsymbol{x}_{t-1}\vert\boldsymbol{x}_{t}, \boldsymbol{x}_{0})$ represents the actual reverse process conditioned on $\bs{x}_0$, while $p_{\theta}(\boldsymbol{x}_{t-1} \vert \boldsymbol{x}_{t})$ denotes the reverse process fitted by our model.
\eqn\ref{eqn:opt-vlb} shows the problem of reconstructing the original data distribution can be transformed into a problem of fitting the reverse process. 
Rewrite $q(\boldsymbol{x}_{t-1}\vert\boldsymbol{x}_{t}, \boldsymbol{x}_{0})$ based on the Bayesian theorem (Appendix~\ref{sec:proof-2}), and we prove it follows a Gaussian distribution over $\bs{x}_{t-1}$:
\begin{equation}
\begin{aligned}
\label{eqn:reverse-prob}
&q(\boldsymbol{x}_{t-1} \vert \boldsymbol{x}_{t}, \boldsymbol{x}_{0}) \sim \mathcal{CN}(\boldsymbol{x}_{t-1}; \tilde{\boldsymbol{\mu}}_{t-1}, \tilde{\boldsymbol{\sigma}}_{t-1}^{2}\mathbf{I}), \\
\tilde{\boldsymbol{\mu}}_{t-1} &=\frac{1}
{\bar{\boldsymbol{\sigma}}_{t}^{2}} (\boldsymbol{\gamma}_{t}\bar{\boldsymbol{\sigma}}_{t-1}^{2} \boldsymbol{x}_t + \bar{\boldsymbol{\gamma}}_{t-1} \boldsymbol{\sigma}^{2}_{t}\boldsymbol{x}_{0}), \ \ \tilde{\boldsymbol{\sigma}}_{t-1} = \frac{\bar{\boldsymbol{\sigma}}_{t-1}}{\bar{\boldsymbol{\sigma}}_{t}} \boldsymbol{\sigma}_{t}.\\
\end{aligned}
\end{equation}
% where $\tilde{\boldsymbol{\mu}}_{t-1} =\frac{1}
% {\bar{\boldsymbol{\sigma}}_{t}^{2}} (\boldsymbol{\gamma}_{t}\bar{\boldsymbol{\sigma}}_{t-1}^{2} \boldsymbol{x}_t + \bar{\boldsymbol{\gamma}}_{t-1} \boldsymbol{\sigma}^{2}_{t}\boldsymbol{x}_{0})$ and $\tilde{\boldsymbol{\sigma}}_{t-1} = \frac{\bar{\boldsymbol{\sigma}}_{t-1}}{\bar{\boldsymbol{\sigma}}_{t}} \boldsymbol{\sigma}_{t}$.
Let's assume that \( p_{\theta}(\boldsymbol{x}_{t-1} \vert \boldsymbol{x}_{t}) \) is a Gaussian Markov process:
\begin{equation}
\label{eqn:reverse-prob-fit}
    p_{\theta}(\boldsymbol{x}_{t-1} \vert \boldsymbol{x}_{t}) \sim \mathcal{CN}(\boldsymbol{x}_{t-1}; \boldsymbol{\mu}_{\theta}(\boldsymbol{x}_{t}), \boldsymbol{\sigma}_{\theta}^{2}(\boldsymbol{x}_{t})  \mathbf{I}).
\end{equation}
% Considering the relationship in \eqn\ref{eqn:forward-0-t}, we have $\boldsymbol{x}_{0} = \frac{\boldsymbol{x}_{t} - \bar{\boldsymbol{\sigma}}_{t}\boldsymbol{\epsilon}}{\bar{\boldsymbol{\gamma}}_{t}}$
% By substituting this into $\tilde{\boldsymbol{\mu}}_{t-1}$, we get:
% \begin{equation}
% \tilde{\boldsymbol{\mu}}_{t-1}=  \frac{1}{\boldsymbol{\gamma}_{t}}(\boldsymbol{x}_{t}  -  \frac{\boldsymbol{\sigma}_{t}}{\bar{\boldsymbol{\sigma}}_{t}}  \boldsymbol{\epsilon}).
% \end{equation}
Therefore, the KL divergence of two Gaussian distributions in \eqn\ref{eqn:opt-vlb} can be simplified as follows:
\begin{equation}
\begin{aligned}
\label{eqn:kl}
& D_{\mathrm{KL}}(q(\boldsymbol{x}_{t-1} \vert \boldsymbol{x}_{t}, \boldsymbol{x}_{0}) \Vert {p_{\theta}(\boldsymbol{x}_{t-1} \vert \boldsymbol{x}_{t})}) \\ 
= & \mathbb{E}_{q(\bs{x}_0)}[\frac{1}{2\tilde{\boldsymbol{\sigma}}_{t}^{2}} \lVert \tilde{\boldsymbol{\mu}}_{t-1} - \boldsymbol{\mu}_{\theta}(\boldsymbol{x}_{t})\rVert^{2} ] + C. \\
% =  & \mathbb{E}_{q(\bs{x}_0)}[\frac{1}{2\tilde{\boldsymbol{\sigma}}_{t}^{2}} \lVert \frac{1}{\boldsymbol{\gamma}_{t}}(\boldsymbol{x}_{t} - \frac{\boldsymbol{\sigma}_{t}}{\bar{\boldsymbol{\sigma}}_{t}} \boldsymbol{\epsilon}) - \boldsymbol{\mu}_{\theta}(\boldsymbol{x}_{t})\rVert^{2} ] + C. 
\end{aligned}
\end{equation}
In summary, the optimization of the the parameterized model $p_{\theta}(\boldsymbol{x}_{t-1} \vert \boldsymbol{x}_{t})$ can be achieved by minimizing the mean square error (MSE) between $\boldsymbol{\mu}_{\theta}$ and $\tilde{\boldsymbol{\mu}}_{t-1}$. In other words, if a model can infer the mean value $\tilde{\boldsymbol{\mu}}_{t-1}$ of the previous step from the input $\bs{x}_t$ of the current diffusion step, then it is competent for the data generation task.

\begin{figure}[t]
    \centering
    \setlength\abovecaptionskip{3pt}
    \includegraphics[]{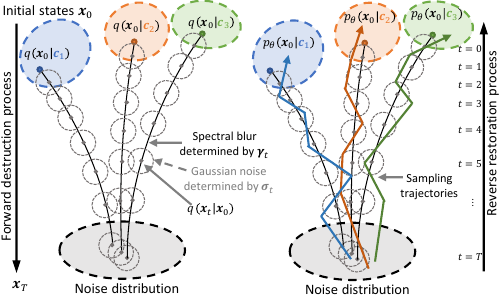}
    \caption{Illustration of the conditional forward and reverse trajectories.}
    \Description{}
    \label{fig:train-sample}
\end{figure}

\subsection{Conditional Generation}
\label{subsec:cond}
% Based the aforementioned analysis, we can optimize the reverse restoration process \( p_{\theta}(\boldsymbol{x}_{t-1} \vert \boldsymbol{x}_{t}) \).
% However, 
In most practical applications, the generation process is expected to be guided by the condition label \( \boldsymbol{c} \), which indicates a specific type of the generated signal (\eg, the signal corresponding to a specific device location or human activity). 

Incorporating the conditional generation mechanism into \sysname offers significant advantages: $(i)$ \textit{Enhanced practicality.} The conditional generation mechanism enables \sysname system to generate signals of different categories based on various condition combinations. This eliminates the need for training separate models for each signal type, significantly improving the model's utility in practical applications. $(ii)$ \textit{Increased signal diversity.} A well-trained conditional generation model creates diverse samples featuring any conceivable combination of characteristics within the condition-label space of the training dataset, which extends the model’s generalizability  beyond the initial scope of the training set, ensuring that data augmentation contributes to performance improvements in downstream tasks.

In this context, the condition input \( \boldsymbol{c} \) defines specific scenarios, including various rooms, Tx-Rx deployments, human activity types, and signal bandwidths. This input guides the generation process to produce data that aligns with the conditional distribution \( p_{\theta}(\bs{x} | \bs{c}) \). An illustration of the conditional forward and reverse processes is presented in \fig\ref{fig:train-sample}.

Following the conclusion of previous work~\cite{sohn2015learning, dhariwal2021diffusion, ho2021classifier}, we directly incorporate the condition \( \boldsymbol{c} \) in both the forward process~\eqn\ref{eqn:forward-prob} and the reverse process~\eqn\ref{eqn:reverse-prob}, and get $q(\boldsymbol{x}_t \vert \boldsymbol{x}_0, \bs{c})$ and $q(\boldsymbol{x}_{t-1} \vert \boldsymbol{x}_{t}, \boldsymbol{x}_{0}, \bs{c})$ respectively. Then, by combining~\eqn\ref{eqn:opt-vlb} and \eqn\ref{eqn:kl}, the optimization can be written as:
\begin{equation}
\begin{aligned}
\label{eqn:opt-cond}
\theta &= \argmin_{\theta}D_{\mathrm{KL}}(q(\boldsymbol{x}_{t-1}\vert\boldsymbol{x}_{t}, \boldsymbol{x}_{0}, \bs{c})\Vert{p_{\theta}(\boldsymbol{x}_{t-1} \vert \boldsymbol{x}_{t}), \bs{c}}) \\
&= \argmin_{\theta} \mathbb{E}_{q(\bs{x}_0)}[ \lVert \tilde{\boldsymbol{\mu}}_{t-1} - \boldsymbol{\mu}_{\theta}(\boldsymbol{x}_{t}(\bs{x}_0, t, \bs{\epsilon}), \bs{c})\rVert^{2} ].
\end{aligned}
\end{equation}

\begin{algorithm}
\caption{\sysname Training.}
\label{alg:training}
\begin{algorithmic}[1]
\Require {Dataset following $\bs{x} \sim q(\bs{x})$ with condition $\bs{c}$}
\Ensure{Trained model $\mu_{\theta}$}
\State Set hyperparameters $T$, $\alpha_t$ and $\bs{g}_{t}$ 
\While{$\mu_{\theta}$ not converged}
\State Sample $\bs{x}_0 \sim q(\bs{x}_0)$ with condition $\bs{c}$ from dataset
\State Sample diffusion step $t \in \mathrm{Uniform}(1, \dots, T)$
\State Sample noise $\bs{\epsilon} \sim \mathcal{CN}(0, \mathbf{I})$
\State Get $\bs{x}_t = \bar{\boldsymbol{\gamma}}_{t}  \boldsymbol{x}_{0} + \bar{\boldsymbol{\sigma}}_{t} \boldsymbol{\epsilon}$ \Comment{\eqn\ref{eqn:forward-0-t}}
\State Calculate $\tilde{\boldsymbol{\mu}}_{t-1}$ based on $\bs{x}_0$ and $\bs{x}_t$ \Comment{\eqn\ref{eqn:reverse-prob}}
\State Minimize $\lVert \tilde{\boldsymbol{\mu}}_{t-1} - \boldsymbol{\mu}_{\theta}(\boldsymbol{x}_{t}(\bs{x}_0, t, \bs{\epsilon}), \bs{c})\rVert^{2}$ \Comment{\eqn\ref{eqn:opt-cond}}
\EndWhile
\end{algorithmic}
\end{algorithm}

\begin{algorithm}
\caption{\sysname Sampling.}
\label{alg:sampling}
\begin{algorithmic}[1]
\Require {Trained model $\mu_{\theta}$, condition $\bs{c}$}
\Ensure{Generated sample $\bs{x}_0$}
\State Set hyperparameters $T$, $\alpha_t$ and $\bs{g}_{t}$ 
\State Sample noise $\bs{\epsilon} \sim \mathcal{CN}(0, \mathbf{I})$
\State Let $\bs{x}_T = \bar{\boldsymbol{\sigma}}_{T}  \boldsymbol{\epsilon}$ \Comment{\eqn\ref{eqn:forward-lim}}
\For{$t = T, \dots, 1$}
\State Get model output $\boldsymbol{\mu}_{\theta}(\boldsymbol{x}_{t}, \bs{c})$, and let $\bs{\sigma}_{\theta} = \tilde{\boldsymbol{\sigma}}_{t-1}$
\State Sample $\bs{x}_{t-1} \sim p_{\theta}(\boldsymbol{x}_{t-1} \vert \boldsymbol{x}_{t})$ with $\boldsymbol{\mu}_{\theta}$ and $\bs{\sigma}_{\theta}$, which means let $\bs{x}_{t-1} =  \boldsymbol{\mu}_{\theta}(\bs{x}_{t}, \bs{c}) + \bs{\sigma}_{\theta} \bs{\epsilon}$
\Comment{\eqn\ref{eqn:reverse-prob-fit}}
\EndFor \\
\Return $\bs{x}_{0}$
\end{algorithmic}
\end{algorithm}

The training process of the parameterized model used for restoration is summarized Algorithm~\ref{alg:training}.
By incorporating the desired signal type as a conditional input, the trained model can iteratively synthesize the original signal from a sampled noise. The generative process is illustrated in Algorithm~\ref{alg:sampling}.

\begin{figure*}[t]
    \centering
    \setlength\abovecaptionskip{5pt}
    \includegraphics[scale=0.78]{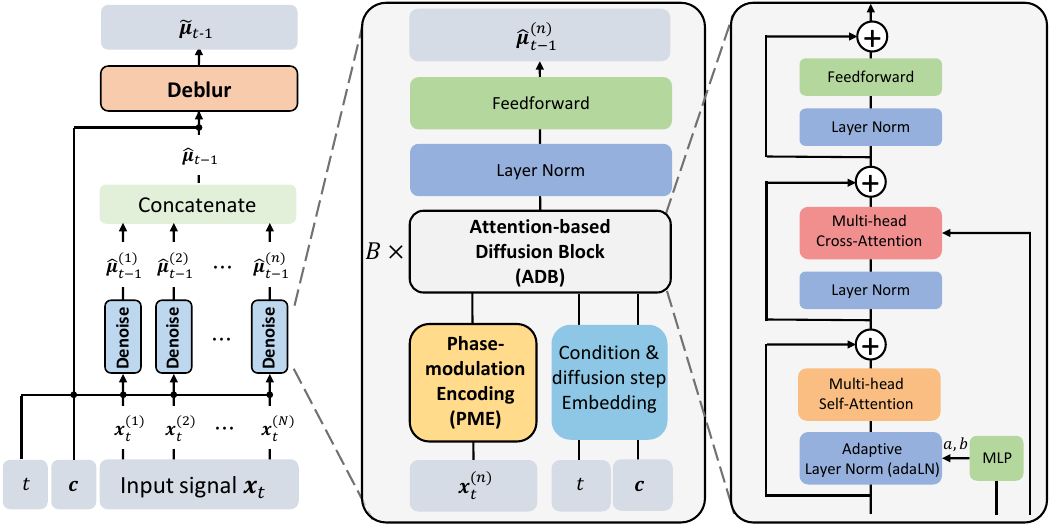}
    \caption{Hierarchical Diffusion Transformer design.}
    \Description{}
    \label{fig:hdt}
\end{figure*}

%% file: contents/4-model.tex
\section{Hierarchical Diffusion Transformer}
\label{sec:model}

To bridge the gap between time-frequency theory and a practical generative model, we introduce a hierarchical diffusion transformer (HDT). 
Our proposed HDT incorporates many innovative designs,  aligning it with the underlying time-frequency diffusion theory and making it adept for RF signal generation. We first introduce the overarching hierarchical design (\secsym\ref{subsec:hierarch}), followed by the detailed design of our proposed attention-based diffusion block (ADB) (\secsym\ref{subsec:block}). Addressing the challenge of complex-valued signal generation, we extend the core design of the classic transformer block~\cite{vaswani2017attention} into the complex domain (\secsym\ref{subsec:complex}). Moreover, we propose phase modulation encoding (PME) (\secsym\ref{subsec:pme}), a novel positional encoding approach tailored for complex-valued neural networks. 

% 为什么选择了层级式的架构？
% 因为tfdiffusion引入的是non-isotropic（非各向同性）的高斯噪声，神经网络难以很好地消除。
% 我们发现了一个机会：每一个采样内部都是isotropic（各向同性）的噪声，各项异性仅仅体现了在时间序列维度的加权上。所以“分而治之”：先独立地为每个样本消除噪声，再在时间序列维度上拟合权重。

\subsection{Hierarchical Architecture}
\label{subsec:hierarch}

From the top perspective, HDT adopts a hierarchical architecture to efficiently decouples the estimation of non-isotropic noise. 
As shown in \fig\ref{fig:hdt}, HDT is divided into two stages: spatial denoising and time-frequency deblurring.

The diffusion step, denoted as $t$, is encoded, thereby informing the model about the current input's diffusion level. The conditional vector $c$ undergoes encoding as well. In conjunction with the input \( \boldsymbol{x}_{t}^{(n)} \), these components engage in computations, striving to discern the latent correlation between the input and its pertinent condition.

Our observation is that the non-isotropic noise can be dissected into two components: 1) Independent Gaussian noise $\bs{\epsilon}$ across both the spatial dimension \(M\) and the temporal dimension \(N\). 2) Different information and noise weights (\ie, $\bar{\gamma}_{t}^{(n)}$ and $\bar{\sigma}_{t}^{(n)}$) along the temporal dimension \(N\).
% Fortunately, the noise added to each sample is independent with the weights.
Therefore, by splitting the time-series data into separate samples, we get $\boldsymbol{x}_t^{(n)} = \bar{\gamma}_{t}^{(n)} \boldsymbol{x}_{0}^{(n)}  + \bar{\sigma}_{t}^{(n)} \boldsymbol{\epsilon}^{(n)}$. Herein, within each sample, both signal and noise weights remain constant. 
Therefore, each spatial denoising module processes a single sample \(\boldsymbol{x}_{t}^{(n)}\) of the input sequence independently. During this stage, denoising circumvents the temporal domain weighting induced by spectral blurring, focusing exclusively on the Gaussian noise $\bs{\epsilon}^{(n)}$ introduced into the original information. This approach resonates with the principles of denoising diffusion~\cite{ho2020denoising}.

Although the spatial denoising module effectively mitigates the impact of noise \(\boldsymbol{\epsilon}\), its individual treatment for each sample disregards the temporal weighting effects originating from spectral blurring. Therefore, the processed results are concatenated as \(\hat{\boldsymbol{\mu}} = [\hat{\boldsymbol{\mu}}^{(1)}, \cdots, \hat{\boldsymbol{\mu}}^{(N)}]\) and serve as sequence input for the time-frequency deblurring module, aiming to estimate the mean value \(\tilde{\boldsymbol{\mu}}_{t-1}\).

% \(\tilde{\boldsymbol{\mu}}_{t}(\boldsymbol{x}_{0},\boldsymbol{\epsilon}, t)\) along the sampling dimension into \(\tilde{\boldsymbol{\mu}}_{t}^{(n)}(\boldsymbol{x}_{0}^{(n)},\boldsymbol{\epsilon}^{(n)}, t)\).

\subsection{Attention-based Diffusion Block}
\label{subsec:block}

As shown in \fig\ref{fig:hdt}, the input data are process by a sequence of transformer blocks in both the denoising and deblur stage. 
We introduce an innovative attention-based diffusion block to jointly analyze the noisy input $\bs{x}_t$, condition $\bs{c}$, and step $t$.

\textbf{Self-attention for feature extraction.} The multi-head self-attention module captures autocorrelation feature from the noisy input and extracts the high-level representations implicit in the signal. Compared to convolutional layers with translation invariance, attention layers are sensitive to the positional information of each sample in the sequence, thus enabling more effective restoration of the original signal.

\textbf{Cross-attention for conditioning.} To enhance the conditional generation capability, \sysname incorporates a cross-attention module to learn the latent associations between the inputs and their corresponding conditions. This module is designed to directly capture the intricate dynamics between the inputs and specified conditions, thereby improving the diversity and fidelity of generated signals. 

\textbf{Adaptive layer normalization for diffusion embedding.} Inspired by the widespread usage of adaptive normalization layer (adaLN)~\cite{perez2018film} in existing conditional generative models~\cite{brock2018large, dhariwal2021diffusion}, we explore replacing standard layer normalization with adaLN. Rather than directly learn dimension-wise scale $a$ and shift parameters $b$, we regress them from the $t$, embedding the diffusion step information into our model. 

\vspace{-5pt}
\subsection{Complex-Valued Module Design}
\label{subsec:complex}
In order to work effectively with complex-valued wireless signals, the \sysname model is designed as a complex-valued neural network. Several adaptations have been made to HDT to facilitate complex-valued operations.

\textbf{Complex-valued attention module.}
Two key improvements have been implemented in the dot-product attention mechanism to accommodate complex number computation: 1) The dot product of the query and key vectors is extended to the hermitian inner product, \(\bs{q}^\mathrm{H}\bs{k}\), which captures the correlation of two vectors in the complex space. This preserves the effective information of both the real and imaginary parts to the fullest extent. 2) Given that the softmax function operates on real numbers, adjustments have been made to make it compatible with complex vectors. Specifically, softmax is applied to the magnitude of the dot product, while the phase information remains unchanged. This modification maintains the probabilistic interpretation of vector relevance. In mathematical terms, the complex-valued attention computation for complex vectors \( \bs{q} \) and \( \bs{k} \) can be expressed as:
\begin{equation}
    \mathrm{softmax}(\lvert \bs{q}^{\ctrans}\bs{k} \rvert) \exp(j\angle{(\bs{q}^{\ctrans}\bs{k})}).
\end{equation}

\textbf{Complex-valued feed-forward module.}
Feed-forward module consists of two main of operations: linear transformation and non-linear activation. A complex-valued linear transformation can be decomposed into real-valued ones~\cite{trabelsi2018deep}. Specifically, for complex-valued input $\bs{x} = \bs{x}_r + j\bs{x}_i$, the transformation with complex weight $\bs{w} = \bs{w}_r + j\bs{w}_i$ and bias $\bs{b} = \bs{b}_r + j\bs{b}_i$ can be written as follows:
\begin{equation}
\begin{aligned}
\bs{w}\bs{x} + \bs{b} = \left[\begin{array}{l}
\Re(\bs{w}\bs{x} + \bs{b}) \\
\Im(\bs{w}\bs{x} + \bs{b})
\end{array}\right]=\left[\begin{array}{rr}
\bs{w}_r& \ -\bs{w}_i \\
\bs{w}_r& \ \bs{w}_i
\end{array}\right] \left[\begin{array}{l}
\bs{x}_r \\
\bs{x}_i
\end{array}\right] + \left[\begin{array}{l}
\bs{b}_r \\
\bs{b}_i
\end{array}\right].
\end{aligned}
\end{equation}
Furthermore, applying an activation function $g(\cdot)$ to a complex value can be seen as activating the real and imaginary parts separately: $g({\bs{x}}) = g({\bs{x}_r}) + jg({\bs{x}_i})$.

\begin{figure}[t]
    \centering
    \setlength\abovecaptionskip{5pt}
    \includegraphics[]{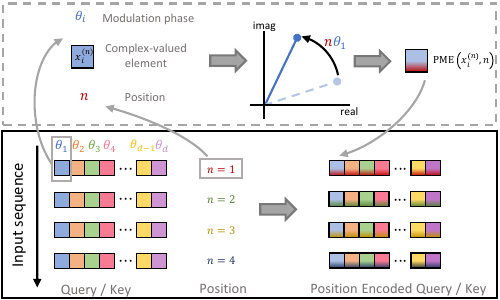}
    \caption{Illustration of phase modulation encoding.}
    \Description{}
    \label{fig:pme}
\end{figure}

\subsection{Phase Modulation Encoding}
\label{subsec:pme}

Leveraging the attention mechanism, a Transformer network parallelly processes the entire sequence. Yet, it lacks inherent capability to discern the positional information of the input. To address this, we introduce an innovative phase modulation encoding (PME) strategy tailored for complex spaces, serving as the positional encoding scheme for HDT.

As illustrated in \fig\ref{fig:pme}, suppose the maximum dimension of each vector in the sequence is \(d\). For the \(i\)-th element of the \(n\)-th vector in the sequence, PME operates as follows:
\begin{equation}
\begin{aligned}
    \mathrm{PME}(\boldsymbol{x}^{(n)}(i), n) &= \boldsymbol{x}_{i}^{(n)} \exp{(jn\theta_i)},
\end{aligned}
\end{equation}
where $\theta_i$ is given by $\theta_i = 10000^{-\frac{i}{d}}$.
This procedure can be conceptualized as a phase modulation process—essentially imparting a specific phase offset to the original data based on the position $n$ in the sequence.

The PME inherently decodes the relative position during computation, establishing its essential role in position encoding. Specifically, when executing the complex-domain Attention operation on the encoded key vector \(\boldsymbol{k}\) and query vector \(\boldsymbol{q}\), it is equivalent to:
\begin{equation}
\label{eqn:pme-relative}
\begin{aligned}
        \mathrm{PME}(\boldsymbol{q}, n)^{\mathrm{H}}\mathrm{PME}(\boldsymbol{k}, m) = \mathrm{PME}(\boldsymbol{q}^{\mathrm{H}}\boldsymbol{k}, m-n).
\end{aligned}
\end{equation}
Therefore, the relative position information \(m-n\) can be derived. This enables our model to learn more proficiently by integrating the positional details of the sequence.

%% file: contents/5-implementation.tex
\section{Implementation}
\label{sec:imp}

We implement \sysname based on PyTorch and train our model on 8 NVIDIA GeForce 3090 GPUs, incorporating essential implementation techniques outlined below.

\noindent\textbf{Exponential moving average.} 
Following common practice in most generative models, we adopt the exponential moving average (EMA) mechanism with a decay rate of $0.999$. EMA calculates a sliding average of the model's weights during training, which improves model robustness. 
% of the trained model.

\noindent\textbf{Weight initialization.} We zero-initialize each final layer before the residual to accelerate large-scale training~\cite{goyal2017accurate}, and apply Xavier uniform initialization~\cite{glorot2010understanding} to other layers, which is a standard weight initialization technique in transformer-based models~\cite{dosovitskiy2021an}.

\noindent\textbf{Hyperparameters.} We train our model using AdamW optimizer~\cite{kinga2015method, loshchilov2018decoupled} with an initial learning rate of $1 \times 10^{-3}$. A step learning rate scheduler with a decay factor of $0.5$ is adopted to improve training efficiency. 
In the training process, we apply a dropout rate of $0.1$ to mitigate overfitting.

\noindent\textbf{Noise scheduling strategy.} In our implementation, the rate of data destruction is designed to increase incrementally from a lower to a higher intensity as the diffusion process progresses. This is aimed at achieving a balance between the model complexity and the generation quality~\cite{kong2020diffwave}. Specifically, we configure the diffusion process with a maximum of \(T = 300\) steps. The noise coefficient, \(\beta_t = \sqrt{1-\alpha_t}\), is set to linearly increase from \(10^{-4}\) to \(0.03\), \ie, \(\beta_t = 10^{-4}t\).
In parallel, the standard deviation of the Gaussian convolution kernel in the frequency domain, denoted as \(\bs{G}_t\), is adjusted to linearly escalate from \(10^{-3}\) to \(0.3\), facilitating the controlled amplification of noise across the diffusion steps.

\noindent\textbf{Data preprocessing.} Each signal sequence from the dataset is either interpolated or downsampled to a consistent length of 512. This guarantees uniformity in the model's input length. Prior to input into our model, each sample within the input sequence is normalized by average signal power, which means each sample is divided by the average L2-norm of all the samples in the sequence.

%% file: contents/6-evaluation.tex
\begin{figure*}[t]
    \centering
    \setlength\abovecaptionskip{0pt}
    \captionsetup[subfloat]{captionskip=0pt}
      \hspace{-1pt}
      \subfloat[Classroom]{
        \includegraphics[scale=0.7]{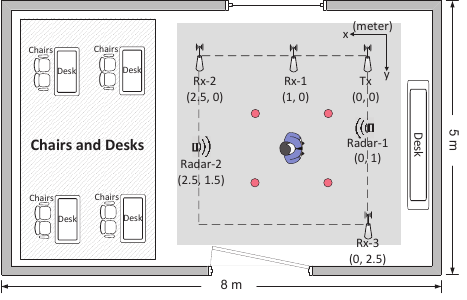}
        \label{fig:exp-scene-1}
      }
      \hspace{-1pt}
      \subfloat[Hall]{
        \includegraphics[scale=0.7]{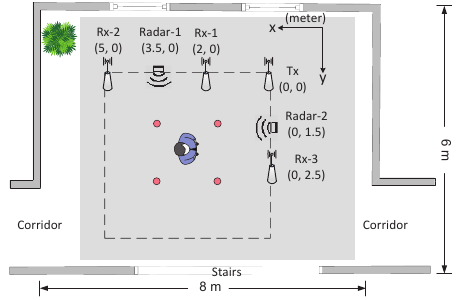}
        \label{fig:exp-scene-2}
      }
      \hspace{-1pt}
      \subfloat[Office]{
        \includegraphics[scale=0.7]{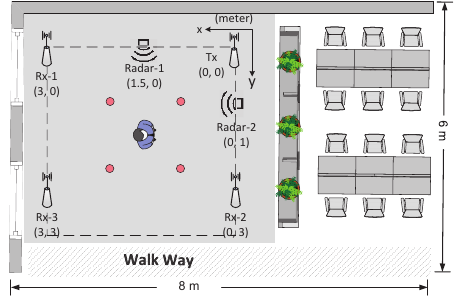}
        \label{fig:exp-scene-3}
      }
    \caption{Experimental scenarios.}
    \Description{}
    \label{fig:exp-scene}
\end{figure*}

% \vspace{-10pt}
\section{Evaluation}
\label{sec:eval}
% In this section, we extensively evaluates signal generation capability of \sysname.

\subsection{Experiment Design}
\label{subsec:exp-design}

\subsubsection{Data Collection.} 

As shown in \fig~\ref{fig:exp-scene}, our dataset comprises wireless signals collected under three distinct scenarios, featuring variations in room selection, device location, and human factors, including their location, orientation, and activities. We compile condition labels for each sequence into a conditional vector $\bs{c}$, guiding both training and sampling phases.
Our research evaluates \sysname's proficiency in producing signals across different modulation modes, focusing on Wi-Fi and FMCW radar signals as two primary types of wireless sensing and communications.

\begin{itemize}
    \item \textbf{Wi-Fi.} We collect Wi-Fi signal based on the commercial NIC IWL5300 working in 5.825 GHz with 40 MHz bandwidth. The transmitter injects Wi-Fi packets to 3 receivers to extract the channel state information (CSI) corresponding to the environment.
    \item \textbf{FMCW.} FMCW signals are recorded using the mmWave radar IWR1443~\cite{iwr1443}. This radar device can be placed at either one of two different locations in each scene, working at a frequency band from 77 GHz to 81 GHz.
\end{itemize}
More than 20,000 Wi-Fi sequences and 13,000 FMCW sequences are collected. Each sequence has an associated condition label indicating the room, device placement, human ID, location, orientation and activity type. All experiments conducted in this paper conform to the IRB policies.

\subsubsection{Comparative Methods.}
\label{subsubsec:comparative}

% 说明为何不和XModal-ID，SynMotion一类对比：需要其他模态辅助，严格来说不属于概率生成模型而只是一个集成的系统;
% 因为目前还没有专门针对无线信号特性设计的（数据驱动的）概率生成模型；
% \sysname专注于通过重建数据分布的方法实现动态、时序数据生成

We compare \sysname with three representative data generation model:
\begin{itemize}
    \item \textbf{DDPM}~\cite{ho2020denoising}. The denoising diffusion probabilistic model (DDPM) introduces Gaussian noise to original data and subsequently learns to reverse this process, thereby generating raw data from the noise.
    \item \textbf{DCGAN}~\cite{radford2015unsupervised}. The deep convolutional generative adversarial network (DCGAN) stands as a widely recognized GAN. In DCGAN, two models (\ie, generator and discriminator) are simultaneously trained in an adversarial manner. Once trained, the generator can produce data that convincingly bypasses the discriminator's scrutiny.
    %We treat the input signal as a 2D complex-valued matrix, and use DCGAN to generate signal given the conditional labels.
    \item \textbf{CVAE}~\cite{NIPS2015_8d55a249}. The conditional variational autoencoder (CVAE) learns the Gaussian implicit representation of the data, thereby enabling data generation. This method is widly adopted in both sensing~\cite{ha2020food} and communication~\cite{liu2021fire} systems to synthesize of wireless features.
\end{itemize}
% 为了能够使以上方案能够处理无线信号数据，我们以复数域神经网络重写了以上模型。
To adapt them for RF signal, we have re-implemented the model using complex-valued neural networks~\cite{trabelsi2018deep}.

\subsubsection{Evaluation Metrics.}
% SSIM 在原始数据层面评估相似性
% FID 在高阶特征层面，评估特征分布的距离度量
% 一定要强调权威性，让人信服或者无从辩驳：目前潜在的攻击点，用STFNets替代Inception，进行信号特征提取是否合理

For a comprehensive evaluation, we adopt two metrics, both of which are commonly used in previous research for evaluating data-driven generative models~\cite{jiang2021focal, nichol2021improved, ho2020denoising, dhariwal2021diffusion, allen-zhu2023forward}. Recognizing that a definitive ``gold standard'' for generative models has not been established, these metrics are among the most authoritative.

\begin{itemize}
\item \textbf{SSIM}~\cite{wang2004image}: The Structural Similarity Index Measure (SSIM) is a prominent criterion for gauging the similarity between two samples by analyzing their means and covariances. We've adapted SSIM for the complex domain, making it suitable for assessing complex-valued signals.
\item \textbf{FID}~\cite{NIPS2017_8a1d6947}: The Fréchet Inception Distance (FID) evaluates generative models by measuring the Fréchet distance between the high-level features of real and synthesized data. We adopt a pretrained STFNets~\cite{yao2019stfnets} as the feature extractor to better fit the property of wireless signals.
\end{itemize}

\subsection{Overall Generation Quality}
\label{subsec:overall}
% \subsubsection{Wi-Fi Signal.}

\begin{figure}[t]
    \centering
    \setlength\abovecaptionskip{-1pt}
    \captionsetup[subfloat]{captionskip=-5pt}
      \subfloat[SSIM]{
        \includegraphics[scale=0.415]{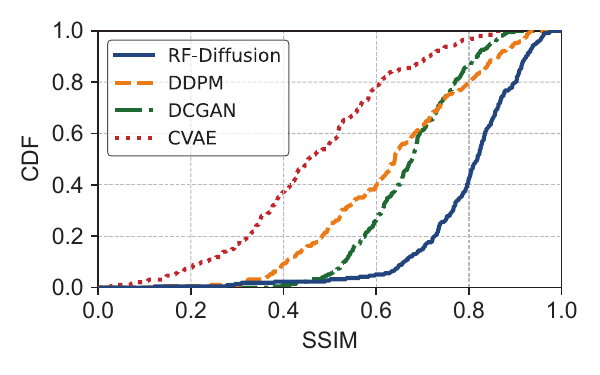}
        \label{fig:exp-overall-wifi-ssim}
      }
      \hspace{-15pt}
      \subfloat[FID]{
        \includegraphics[scale=0.415]{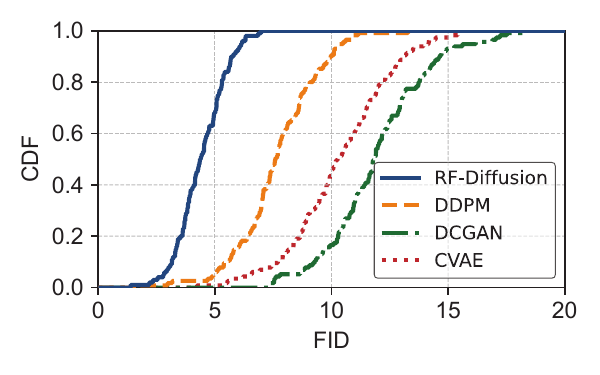}
        \label{fig:exp-overall-wifi-fid}
      }
    \caption{Wi-Fi signal generation quality.}
    \Description{}
    \label{fig:exp-overall-wifi}
\end{figure}

\begin{figure}[t]
    \centering
    \setlength\abovecaptionskip{-1pt}
    \captionsetup[subfloat]{captionskip=-5pt}
      \subfloat[SSIM]{
        \includegraphics[scale=0.4215]{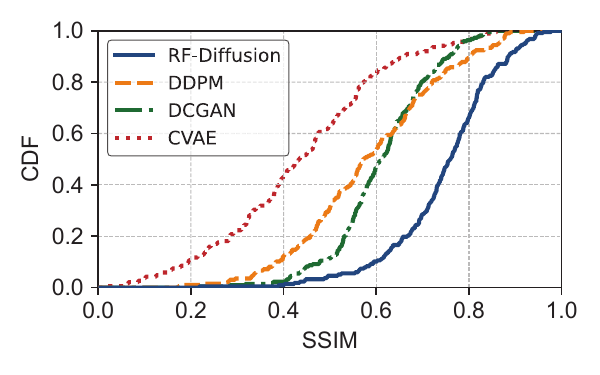}
        \label{fig:exp-overall-fmcw-ssim}
      }
      \hspace{-15pt}
      \subfloat[FID]{
        \includegraphics[scale=0.4215]{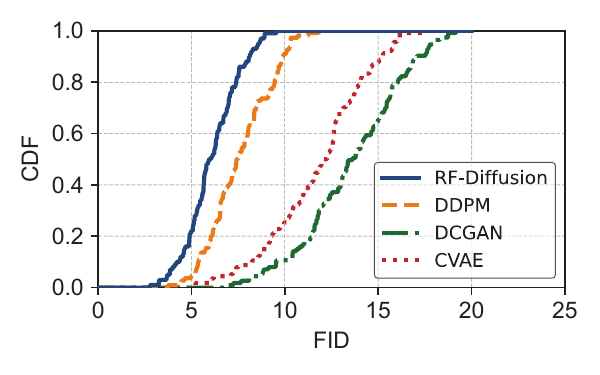}
        \label{fig:exp-overall-fmcw-fid}
      }
    \caption{FMCW signal generation quality.}
    \Description{}
    \label{fig:exp-overall-fmcw}
\end{figure}

The evaluation result for \sysname on Wi-Fi and FMCW signal are illustrated in \fig\ref{fig:exp-overall-wifi} and \fig\ref{fig:exp-overall-fmcw} respectively. As shown, our proposed \sysname has proved the superiority over comparative methods on both two metrics.

Specifically, as shown in \fig\ref{fig:exp-overall-wifi}, \sysname generates Wi-Fi signal with an average SSIM of $0.81$, exceeding DDPM, DCGAN, and CVAE by $25.4\%$, $18.6\%$ and $71.3\%$ respectively. \sysname achieves an FID of $4.42$, outperforming the above comparative methods by $42.4\%$, $63.0\%$, and $57.3\%$.

\sysname also outperforms the comparative methods in terms of generating high-fidelity FMCW signals. As shown in \fig\ref{fig:exp-overall-fmcw}, the FMCW signal generated by \sysname achieves an average SSIM of $0.75$ and an average FID of $6.10$.

The impressive performance of \sysname can be attributed to several key factors: 1) Our proposed time-frequency diffusion adopted by \sysname emphasizes refining the frequency spectrum of the RF signal, thereby preserving finer spectral details in the generated signals, which is difficult to be captured by other methods. 2) Through its iterative generation approach, \sysname attains precise reconstruction of data details via multi-step approximations, leading to a superior quality of generated data. 3) In contrast to DCGAN, which optimizes two models concurrently, \sysname's loss function is more streamlined and its training process more stable, ensuring a richer diversity in the generated signal and contributing to a commendable FID score.

% As shown in \fig\ref{fig:exp-overall-wifi}, \sysname generates high-fidelity Wi-Fi signal with an average SSIM of $0.79$, exceeding DDPM, DCGAN, and CVAE by $25.4\%$, $18.6\%$ and $71.3\%$ respectively. As for the FID, \sysname achieves $4.42$, outperforming the above comparative methods by $42.4\%$, $63.0\%$, and $57.3\%$ respectively.

% % \subsubsection{FMCW Signal.} 

% \sysname also outperforms the comparative methods in terms of generating FMCW signals. As shown in \fig\ref{fig:exp-overall-fmcw}, \sysname achieves an average SSIM of $0.75$ and FID of $6.10$.

\begin{figure*}[t]
    \centering
    \setlength\abovecaptionskip{0pt}
    \begin{minipage}[t]{0.33\textwidth}\centering
      \setlength\abovecaptionskip{-5pt}
      \includegraphics[width=\fullfigwidth]{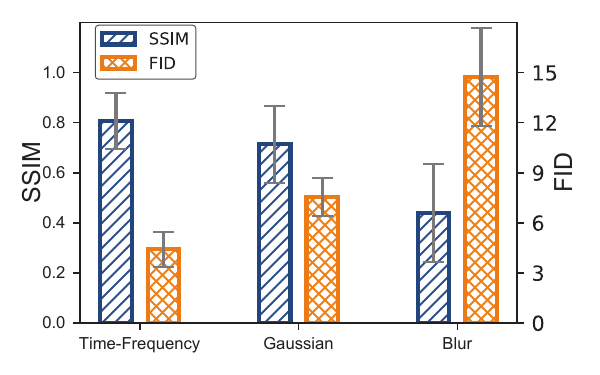}
        \caption{Impact of diffusion method.}
        \label{fig:exp-micro-theory}
    \end{minipage}
    \hspace{-0.1in}
    \begin{minipage}[t]{0.33\textwidth}\centering
      \setlength\abovecaptionskip{-5pt}
      \includegraphics[width=\fullfigwidth]{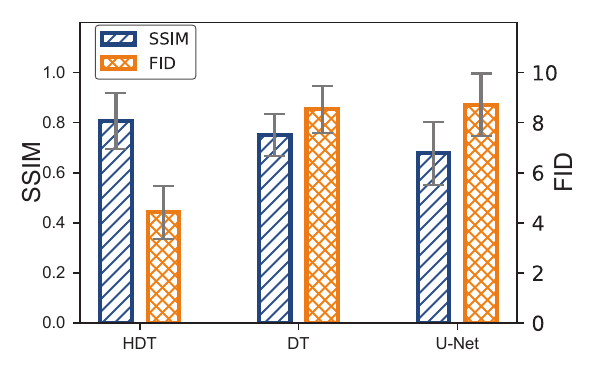}
        \caption{Impact of network design.}
        \label{fig:exp-micro-model}
    \end{minipage}
    \hspace{-0.1in}
    \begin{minipage}[t]{0.33\textwidth}\centering
        \setlength\abovecaptionskip{-5pt}
        \includegraphics[width=\fullfigwidth]{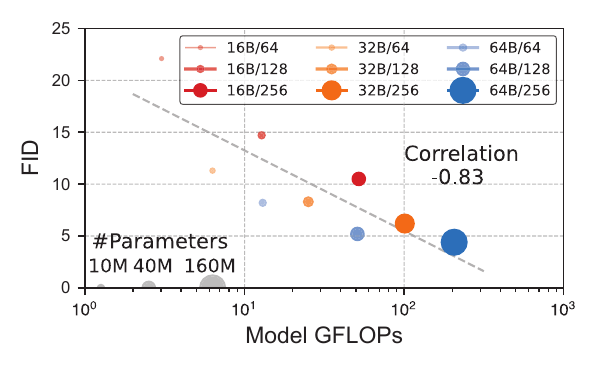}
        \caption{Scalability analysis.}
      \label{fig:exp-scalability}
    \end{minipage}
\end{figure*}

\vspace{-7pt}
\subsection{Micro-benchmarks}

\subsubsection{Impact of Diffusion Methods.} To validate the efficacy of our proposed time-frequency diffusion theory, we retained the network model architecture of \sysname but replaced the time-frequency diffusion process with two alternate schemes: 1) Gaussian diffusion, which is similar to DDPM and by only introduces Gaussian noise to the signal amplitude, and 2) blur diffusion which only performs spectral blurring. 
As depicted in \fig\ref{fig:exp-micro-theory}, our time-frequency diffusion theory consistently outperforms both in terms of the SSIM and FID metrics. Specifically, the SSIM values for time-frequency diffusion, Gaussian diffusion, and blur diffusion stand at $0.81$, $0.71$, and $0.45$, respectively. This translates to the time-frequency diffusion offering an SSIM improvement of $13.9\%$ over Gaussian diffusion and a notable $79.2\%$ over blur diffusion. In terms of the FID, the time-frequency diffusion surpasses the other two methods by margins of $41.3\%$ and $83.5\%$, respectively. The results indicates that the time-frequency diffusion theory successfully incorporates two diffusion methods on orthogonal spaces, and thus achieving complementary benefits.

\subsubsection{Impact of Network Design.} To demonstrate the advantages of our proposed hierarchical diffusion transformer (HDT), we compare it against: 1) single-stage diffusion transformer (SDT), which is a simplified form of HDT, with only one stage for end-to-end data restoration, and 2) U-Net~\cite{ronneberger2015u}, a popular choice in prevalent diffusion models. 
As shown in \fig\ref{fig:exp-micro-model}, our proposed HDT outperforms the SDT and U-Net. Specifically, the SSIM for HDT, SDT, and U-Net are $0.81$, $0.75$, and $0.68$ respectively. This indicate that HDT achieves a SSIM boost of $7.7\%$ over SDT and a significant $18.9\%$ increment compared to U-Net. When assessed using the FID metric, HDT continues to lead by margins of $48.2\%$ and $49.3\%$ against SDT and U-Net, respectively.
The outstanding performance benefits from the follows aspects: 1) Compared with SDT, HDT can efficiently decouple the non-isotropic noise introduced in the diffusion process and eliminate it through two sequential stages; 2) Compared with translation-invariant U-Net, HDT's transformer architecture can effectively distinguish the signal characteristics at different times, thereby achieving more accurate signal generation.

\subsubsection{Scalability Analysis.} Scalability refers to a model's ability to enhance its performance with increasing size, which is critical for large generative models like \sysname. To verify the scalability of \sysname, we trained 9 models of different sizes, exploring different numbers of attention-based diffusion blocks (16B, 32B, 64B) and hidden dimensions (64, 128, 256). \fig\ref{fig:exp-scalability} illustrates that the FID performance of \sysname is strongly correlated with model parameters and GFLOPs, indicating that scaling up model parameters and additional model computation is the critical ingredient for better performance. Increasing the model size is anticipated to further enhance \sysname's performance.

% Comparative Methods

% 我们将所提出的系统与最新的几类不同的数据生成解决方案作对比。
% 现有的一些数据合成方案，需要借助其他模态并针对指定类型的特征进行，因此并不适合被作为对比方案。

%% file: contents/7-case-study.tex
\section{Case Study}
\label{sec:case-study}
% 本节中，我们通过无线感知与信道预测两个常见的下游任务，展示系统在无线感知与通信系统中的应用潜力。
This section showcases how \sysname benefits wireless researches in two distinct downstream tasks: Wi-Fi-based gesture recognition and 5G FDD channel estimation.

\subsection{Wi-Fi Gesture Recognition}
\label{subsec:case-sensing}
% 作为一个生成式模型，系统可以作为一个数据增强器插件，（在不对原有模型进行内部修改的情况下）提升现有无线感知系统的性能。
% As a generative model, \sysname can be leveraged as a data augmenter to boost the performance of existing wireless sensing systems, without making any modifications to original model structure.
Wireless sensing~\cite{chi2022wi, yang2023slnet, zhang2023push, gao2023wi} has emerged as a major research focus. By serving as a data augmentor, \sysname can boost the performance of existing wireless sensing systems, all while preserving the original model structure without any modifications.
% Specifically, we first train \sysname with the real-world dataset. Once trained, \sysname synthesizes the specified type of wireless signal based on the condition label. Synthetic data will be mixed with the real-world dataset and used together for the training of the wireless sensing model. 
In particular, our approach involves initially training \sysname using a real-world dataset. Subsequently, \sysname generates synthetic RF signals of the designated type, guided by condition labels. These synthetic samples are then integrated with the original dataset, collectively employed to train the wireless sensing model. 
Both \sysname-augmented solution and baseline are fundamentally based on the same real-world dataset, ensuring a fair comparison, as \sysname itself is trained on this real-world dataset and no extra data is ever involved. 

We illustrate this approach through the case of Wi-Fi-based gesture recognition and evaluate the performance gains achieved by integrating \sysname into established gesture recognition models.

\begin{figure}[t]
    \centering
    \setlength\abovecaptionskip{0pt}
    \captionsetup[subfloat]{captionskip=-5pt}
      \subfloat[Cross-domain]{
        \includegraphics[scale=0.416]{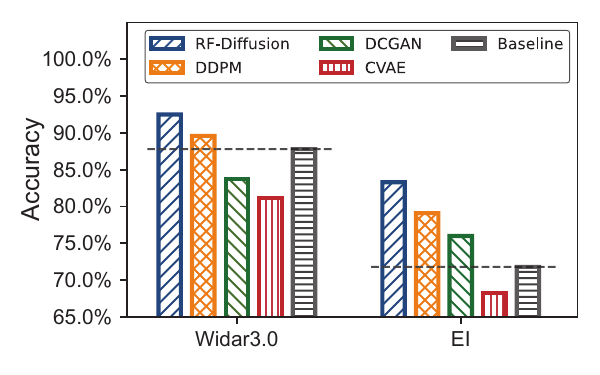}
        \label{fig:exp-sensing-cross}
      }
      \hspace{-12pt}
      \subfloat[In-domain]{
        \includegraphics[scale=0.416]{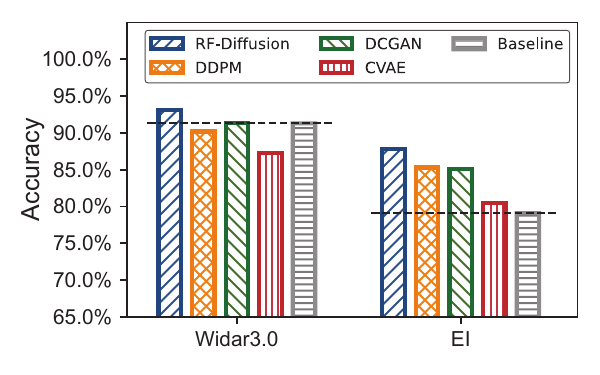}
        \label{fig:exp-sensing-in}
      }
    \caption{Performance of augmented Wi-Fi sensing.}
    \Description{}
    \label{fig:exp-sensing}
\end{figure}

\subsubsection{Experiment Design.}

We select two different types of Wi-Fi-based model for a comprehensive evaluation:
\begin{itemize}
    \item \textbf{Widar3.0}~\cite{zheng2019zero} is a gesture recognition model founded on physical principles. It initially extracts features from raw signals and subsequently conducts recognition through a deep neural network.
    \item \textbf{EI}~\cite{jiang2018towards} is a data-driven end-to-end human activity recognition model that takes raw signal as input.
\end{itemize}

We utilize the publicly available dataset from Widar3.0~\cite{zheng2019zero} to assess performance. This evaluation encompasses scenarios where \sysname and comparative methods (\secsym\ref{subsubsec:comparative}) were employed as data augmentors.

\begin{figure}[t]
    \centering
    \setlength\abovecaptionskip{-5pt}
    \includegraphics[scale=0.42]{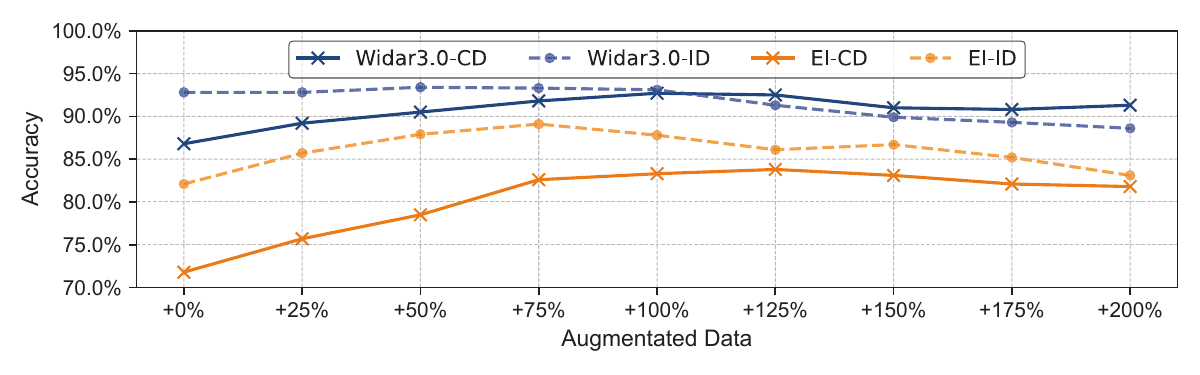}
    \caption{Impact of synthetic data volume.}
    \label{fig:exp-sensing-data}
\end{figure}

\subsubsection{Cross-domain Evaluation.}

We first evaluate the sensing performance when the training and testing set are from different domains (\ie, room, device placement, human location, orientation, \etc), a common scenario in real-world wireless sensing system deployments.
We synthesize an equivalent volume of data as the real-world dataset using the pre-trained \sysname. Subsequently, both synthesized and authentic datasets are used for training.
As shown in \fig\ref{fig:exp-sensing-cross}, integrating \sysname brings performance improvements of $4.7\%$ and $11.5\%$ for Widar3.0 and EI, respectively. 
Integrating DDPM can bring relatively limited performance gains of only $1.8\%$ and $7.3\%$, respectively. Additionally, the integration of DCGAN or CVAE may result in a degradation of recognition accuracy due to deviations in the synthetic data distribution from the original data distribution.
  
Compared with Widar 3.0, the EI model obtains a more significant improvement since: 1) EI is more sensitive to the data volume and data diversity as an end-to-end DNN; 2) the information in the wireless signals generated in a data-driven manner cannot be fully exploited in Widar3.0 when being converted into physical features.

% Compared with to Widar 3.0, the EI model achieves a more substantial performance improvement for two primary reasons: 1) being an end-to-end deep learning model, EI exhibits heightened responsiveness to variations in data volume and data diversity; 2) the information in the wireless signals generated in a data-driven manner cannot be fully exploited in Widar3.0 when being converted into physical features.
 
In conclusion, \sysname enhances the cross-domain performance of wireless sensing systems in two aspects:
\begin{itemize}
    \item \textbf{Enhanced data diversity.} Synthetic training data with higher diversity avoid model overfitting and thus implicitly improves the model's domain generalization ability.
    \item \textbf{Feature distillation.} The generative model \sysname implicitly imparts its learned signal features to the recognition model through synthetic training data, contributing to improved performance.
\end{itemize}

\subsubsection{In-domain Evaluation.}

In the in-domain scenario, the training and testing data are from the same domain. As shown in \fig\ref{fig:exp-sensing-in}, the integration of \sysname yields performance improvements of $1.8\%$ and $8.7\%$ for Widar3.0 and EI, respectively. 
Overall, compared with the cross-domain scenario, the performance gain in the in-domain case is relatively modest. This is attributed to the limited impact of diverse synthetic training data on enhancing the model's performance within the same domain.
For in-domain testing, even with a less diverse synthetic data generated by DCGAN, an obvious performance gain can be achieved.

\subsubsection{Impact of Synthesized Data Ratio.}

We further investigate the impact of the synthesized data ratio used for training to provide more insights. 
We evaluate Widar 3.0 and EI in both cross-domain (CD) and in-domain (ID) cases.

As shown in the \fig\ref{fig:exp-sensing-data}, we introduce varying quantities of synthetic data (from $+25\%$ to $+200\%$) to the real-world dataset for joint training of the recognition model. Notably, as the volume of synthetic data increases, the trend in recognition accuracy exhibits an ascent to a peak followed by a decline.
Specifically, in the cross-domain case, Widar3.0 reaches the highest accuracy of $92.7\%$ with $+100\%$ synthetic data, while EI reaches the highest accuracy of $83.8\%$ with $+125\%$ synthetic data.
In the in-domain case, Widar3.0 achieved the highest accuracy of $93.4\%$ with $+50\%$ synthetic data, while EI achieved the highest accuracy of $89.1\%$ with $+75\%$ synthetic data.
Drawing from these statistical findings, we deduce the following insights: 1) For most wireless recognition models, judicious incorporation of synthetic data into the training set can effectively enhance model performance. 2) Excessive introduction of synthetic data can potentially shift the training data distribution away from the original, consequently diminishing recognition accuracy. 3) The cross-domain scenario requires a greater infusion of synthetic data into the training set to achieve optimal model performance compared to the in-domain scenario. 4) Data-driven end-to-end models (\eg, EI) reap more substantial benefits from data augmentation facilitated by \sysname.

\subsection{5G FDD Channel Estimation}
\label{subsec:case-channel}

\begin{figure}[t]
    \centering
    \setlength\abovecaptionskip{0pt}
    \captionsetup[subfloat]{captionskip=-5pt}
      \subfloat[Channel amplitude and phase]{
        \includegraphics[scale=0.4]{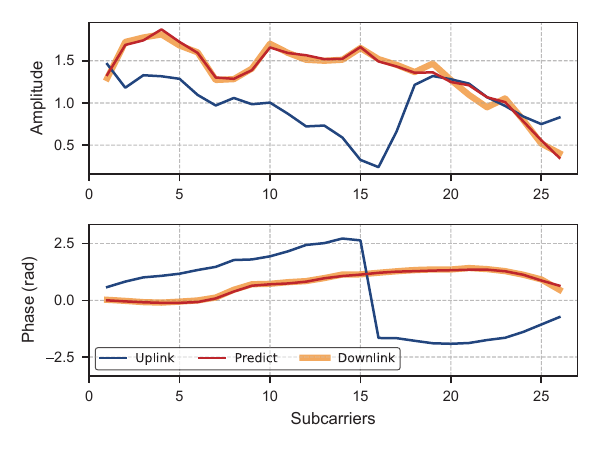}
        \label{fig:exp-channel-sample}
      }
      \hspace{-5pt}
      \subfloat[SNR]{
        \includegraphics[scale=0.4]{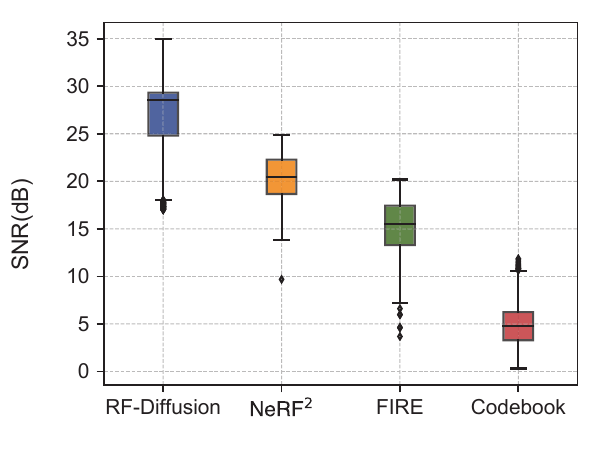}
        \label{fig:exp-channel-snr}
      }
    \caption{Performance of channel estimation.}
    \Description{}
    \label{fig:exp-channel}
\end{figure}

In this section, we discuss how \sysname enables channel estimation of the Frequency Domain Duplex (FDD) system in 5G, where the uplink and downlink transmissions operate at different frequency bands. Therefore, the principle of reciprocity that two link channels are equal no longer holds~\cite{liu2021fire}.
To estimate the downlink channel state, client devices must receive additional symbols from a base station with a massive antenna array and send back the estimated results, causing unsustainable overheads.
To solve this problem, substantial research is devoted to predicting the downlink channel by observing the uplink channel state information. For example, FNN~\cite{bakshi2019fast} and FIRE~\cite{liu2021fire} make use of a fully connected network and a VAE to transfer the estimated CSI from the uplink to the downlink, respectively.

We discover that by employing the uplink CSI as conditional input, \sysname demonstrates the capacity to estimate downlink channel CSI in a generative manner.
Specifically, in \sysname, the downlink CSI $\bs{x}_\mathrm{down}$ serves as the target data for generation, while the uplink CSI is encoded as the condition $\bs{c}_\mathrm{up}$ and input into the model. 
The trained \sysname learns the correlation between $\bs{c}_\mathrm{up}$ and $\bs{x}_\mathrm{down}$, thereby accomplishing the channel estimation task.
This efficacy is rooted in the assumption of shared propagation paths, positing that both link channels are shaped by the same underlying physical environment~\cite{huang2019deep, vasisht2016eliminating}.

\subsubsection{Experiment Design.}

Our evaluation is based on the publicly available dataset Argos~\cite{shepard2016understanding}, which is a real-world MIMO dataset collected in a complex environment with a large number of non-line-of-sight (NLoS) propagation. Each CSI frame contains 52 subcarriers. 
Similar to previous works~\cite{zhao2023nerf, liu2021fire}, we designate the initial 26 subcarriers for the uplink channel, while the remaining 26 are allocated to the downlink channel.
For a comprehensive evaluation, we compare our approach against three different types of channel estimation solutions:
\begin{itemize}
    \item \textbf{NeRF}$^{2}$~\cite{zhao2023nerf} implicitly learns the signal propagation environment from the uplink channel state based on a neural network, and then estimate the downlink channel.
    \item \textbf{FIRE}~\cite{liu2021fire} is a channel estimation system based on VAE, which compresses the uplink channel CSI into a latent space representation and further transforms it into a downlink channel estimation.
    \item \textbf{Codebook}~\cite{kaltenberger2008performance}, commonly used in standard implementations per the 3GPP physical channel standard\cite{3gpp}, requires both base stations and clients to maintain a codebook of vectors created using predefined rules. Clients measure the channel locally, select the closest codebook vectors, and send the corresponding indices back to the base station.
\end{itemize}

\subsubsection{Channel Estimation Accuracy.}

As shown in \fig\ref{fig:exp-channel-sample}, when we input the blue uplink CSI as a condition into the trained \sysname, the red downlink estimate will be output, which closely aligns with the ground truth downlink channel state.
Assessment of channel estimation accuracy employs the Signal-to-Noise Ratio (SNR) metric~\cite{zhao2023nerf, liu2021fire}. This metric gauges the congruence between the estimated downlink channel $\boldsymbol{x}_{\mathrm{est}}$ and the ground truth $\boldsymbol{x}_{\mathrm{down}}$ through the following formulation:
\begin{equation}
\mathrm{SNR} = -10\log_{10}\left( \frac{\lVert \bs{x}_{\mathrm{down}} - \bs{x}_{\mathrm{est}} \rVert^{2}}{\lVert \bs{x}_{\mathrm{down}} \rVert^{2}}\right).
\end{equation}
A higher positive SNR corresponds to enhanced proximity between the predicted channel and the ground truth.
As shown in \fig\ref{fig:exp-channel-snr}, \sysname achieves the highest SNR among all comparative methods with an average SNR of $27.01$, outperforming NeRF$^{2}$ and FIRE by $34.6\%$ and $77.5\%$ respectively, and achieves more than $5\times$ performance gain compared with the standard implementation based on codebook.

The observed underperformance of NeRF$^{2}$ can be attributed to its treatment of the signal propagation space as a time-invariant system, a characterization that may not hold in practical scenarios. VAE-based FIRE and codebook-based methods fall short in the fine-grained characterization of the underlying distribution of channel states. In contrast, \sysname adeptly learns the intricate correlation between the uplink and downlink channels, leveraging its robust modeling capacity to achieve highly accurate channel estimation.

%% file: contents/8-related-work.tex
\section{Related Work}
\label{sec:related-work}

We briefly review the related works in the following.

\textbf{Diffusion probabilistic models.} 
Diffusion probabilistic models~\cite{yang2022diffusion,sohl2015deep} have emerged as a powerful new family of deep generative models with record-breaking performance in many applications~\cite{dhariwal2021diffusion}, including image synthesis, point cloud completion, and natural language processing, etc. 
One of the best-known diffusion model is the DDPM~\cite{sohl2015deep}, which progressively destruct data by injecting gaussian noise, then learn to reverse this process for high-fidelity sample generation. On this basis, DDIM~\cite{nichol2021improved} expedites reverse sampling, while LDM~\cite{rombach2022high} conducts diffusion in latent space to curtail computational overhead. The above schemes have been widely used in a wide range of tasks such as image super-resolution~\cite{saharia2022image, ho2022cascaded}, inpainting~\cite{lugmayr2022repaint}, and style transfer~\cite{saharia2022palette}. The most recent studies~\cite{lee2022progressive, rissanen2023generative} have successfully applied a combination of blurring and additive noise to the image, yielding satisfactory results.
Although first proposed for image generation, the diffusion model's versatility extends to other domains including point cloud completion~\cite{zhou20213d, lyu2021conditional}, text generation~\cite{austin2021structured, gong2023diffuseq}, audio synthesis~\cite{kong2020diffwave, chen2021wavegrad}, and beyond. 
In addition, diffusion model has a great potential for multi-modal generation. By integrating pre-trained language model~\cite{radford2021learning}, the diffusion models achieve impressive performance in text-to-image~\cite{pmlr-v162-nichol22a, ramesh2022hierarchical} and  text-to-audio~\cite{popov2021grad} tasks. 

\sysname, in contrast, stands as the pioneering diffusion model tailored for wireless signal generation. It introduces an innovative time-frequency diffusion process, which regulates noise and blurring across two orthogonal domains, thus encompassing both temporal and spectral intricacies of wireless signals. By generating high-fidelity signals, \sysname benefits wireless applications like Wi-Fi sensing and 5G channel estimation.

\textbf{Signal generation in wireless systems.} 
Conventional wireless signal generation schemes are mainly based on modeling and simulation.
In particular, these methods involve utilizing LiDAR-scanned 3D models, and employing electromagnetic (EM) ray tracing techniques~\cite{mckown1991ray} to simulate the distribution of wireless signals.
Recent studies~\cite{korany2019xmodal, cai2020teaching, zhang2022synthesized} have integrated vision-based human reconstruction techniques with signal propagation models, enabling the generation of wireless signals that interact with the human body. Unfortunately, the above schemes fails to model the structure material and physical characteristics, which constraints their performance in real-world applications.
% Additionally, obtaining a 3D model with accuracy compatible with RF signal wavelengths remains a challenge and will significantly raise the system expenses.
The recently proposed NeRF$^2$~\cite{zhao2023nerf} learns the properties of the signal propagation space based on a deep neural network and then accomplishes the signal generation task. However, NeRF$^2$ is limited to specific static scenarios and degrades for dynamic real-world scenarios.
RF-EATS~\cite{ha2020food} and FallDar~\cite{yang2022rethinking} employ Variational Autoencoders (VAEs) to extract environment-independent features, thereby enhancing the generalizability of wireless sensing models. Additionally, other studies have utilized Generative Adversarial Networks (GANs) to generate Doppler spectrum~\cite{erol2019gan}. Other research endeavors have addressed channel estimation in wireless communication systems using either GANs~\cite{doshi2022over, balevi2021wideband} or VAEs~\cite{baur2022variational, liu2021fire}.
Nonetheless, due to their limited representation capabilities, solutions based on GANs and VAEs struggle to faithfully characterize the intrinsic properties of original wireless signals. Consequently, the aforementioned systems are suitable solely for specific tasks, lacking the competence for general-purpose wireless data generation.

In contrast, \sysname, as a versatile generative model for wireless signals, can proficiently generate fine-grained signals in high-fidelity, even within dynamic scenarios.

%% file: contents/9-discussion.tex
\vspace{-5pt}
\section{Discussion and Future Work}
\label{sec:discussion}

\sysname is a pioneering attempt towards diffusion-based RF signal generation, and there is room for continued research in various perspectives.

\noindent $\bullet$ \textbf{\sysname for data-driven downstream tasks.} Extensive practices~\cite{he2023is, shivashankar2023semantic, azizi2023synthetic,zheng2023toward} indicate that synthetic data from generative models significantly enhances data-driven downstream tasks.
As a conditional generative model, \sysname effectively captures the representative features and their novel combinations, while randomizing non-essential details. This approach allows for the generation of innovative data samples that extend beyond the initial scope of the dataset, thus improving the generalization ability of downstream models.
This paper specifically explores and experiments with applying \sysname to augment Wi-Fi gesture recognition, demonstrating its potential. However, the applicability of \sysname extends to any data-driven task in wireless communication and sensing. 

\noindent $\bullet$ \textbf{\sysname as a simulation tool.} As a probabilistic generative model, \sysname operates independently of any signal propagation assumptions and does not require pre-modeling of the environment. This flexibility implies that, while \sysname offers novel opportunities for signal synthesis, it may not achieve the same level of stability and precision as traditional signal simulation tools in all scenarios. \sysname is not designed to supplant simulation tools but rather to introduce a novel data-driven approach for signal synthesis, which is particularly valuable in complex and dynamic environments, such as indoor spaces with human activity, where accurate modeling poses challenges.

\noindent $\bullet$ \textbf{Autoregressive signal generation.} 
\sysname, a non-autoregressive generative model, processes time series as a unified entity, necessitating downsampling and interpolation for variable-length sequences, which limits its versatility. The advent of autoregressive models like GPT introduces alternative methods for time-series signal generation, which improves adaptability for sequences of differing lengths and enable effective exploration of temporal correlation features.

%% file: contents/10-conclusion.tex
\section{Conclusion}
\label{sec:conclusion}

This paper presents \sysname, the pioneering generative diffusion model designed for RF signals. \sysname excels in generating high-fidelity time-series signals by employing a novel time-frequency diffusion process. This process captures the intricate characteristics of RF signals across spatial, temporal, and frequency domains. This theoretical framework is then translated into a practical generative model based on the hierarchical diffusion transformer. \sysname exhibits remarkable versatility. It holds significant potential for essential wireless tasks, ranging from boosting the accuracy of wireless sensing systems, to estimating channel states in communication systems, shedding light on the applications of AIGC in wireless research.

%% file: contents/appendix.tex
\appendix
\section{Convergence of Forward Destruction Process}
\label{sec:proof-1}

As \( T\to\infty \), the forward process converges to a distribution independent of the original signal. The above proposition is equivalent to the following two conditions: (1) \(\lim_{T \to \infty} \bar{\boldsymbol{\mu}}_{T} = \mb{0} \), (2) \(\lim_{T \to \infty} \bar{\boldsymbol{\sigma}}_{T} < \infty\).
We find a sufficient condition for the above to hold true: all element in \( \boldsymbol{\gamma}_t = \sqrt{\alpha_t}\boldsymbol{g}_{t} \) should be less than 1, \ie, \( \forall n, \gamma_{t}^{(n)} < 1 \). Under this condition, according to \eqn\ref{eqn:forward-0-t}, \( \lim_{T \to \infty} \bar{\boldsymbol{\mu}}_{T} = \lim_{T \to \infty} \bar{\boldsymbol{\gamma}}_{T} \bs{x}_0 = \mb{0} \) holds.
% A sufficient condition to ensure is: all element in \( \boldsymbol{\gamma}_t = \sqrt{\alpha_t}\boldsymbol{g}_{t} \) should be less than 1, \ie, \( \forall n, \gamma_{t}^{(n)} < 1 \). Under this condition, according to \eqn\ref{eqn:forward_0_t}, \( \lim_{T \to \infty} \bar{\boldsymbol{\mu}}_{T} = \lim_{T \to \infty} \bar{\boldsymbol{\gamma}}_{T} \bs{x}_0 = \mb{0} \) holds.
Let \( \alpha_\mathrm{min} = \min(\alpha_t), t \in [1, T] \) and \( {\gamma}^{(n)}_{\mathrm{max}} = \max(\gamma_{t}^{(n)}) \) and \( \boldsymbol{\gamma}_{\mathrm{max}} = ({\gamma}^{(1)}_{\mathrm{max}}, \dots, {\gamma}^{(N)}_{\mathrm{max}}) \). It can be proven that:
\begin{equation}
\begin{aligned}
& \lim_{T \to \infty} \bar{\boldsymbol{\sigma}}_{T} = \lim_{T \to \infty} \sum_{t = 1}^{T} {(\sqrt{1 - \alpha_t} \frac{\bar{\boldsymbol{\gamma}}_{T}}{\bar{\boldsymbol{\gamma}}_{t}})} \\
& \leq \sqrt{1 - \alpha_{\mathrm{min}}} \lim_{T \to \infty} \sum_{t = 1}^{T} {(\boldsymbol{\gamma}_{\mathrm{max}})^{t-1}} = \frac{\sqrt{1 - \alpha_{\mathrm{min}}}}{1 - \boldsymbol{\gamma}_{\mathrm{max}}} < \infty
\end{aligned}
\end{equation}

\section{Reverse Process Distribution}
\label{sec:proof-2}
Based on the Bayes' theorem, we get:
\begin{equation}
\begin{aligned}
&  q(\boldsymbol{x}_{t-1}\vert\boldsymbol{x}_{t}, \boldsymbol{x}_{0}) =q(\boldsymbol{x}_{t}\vert\boldsymbol{x}_{t-1}, \boldsymbol{x}_{0})\frac{q(\boldsymbol{x}_{t-1} \vert \boldsymbol{x}_0)}{q(\boldsymbol{x}_{t} \vert \boldsymbol{x}_{0})}\\
\propto&\exp(-\frac{1}{2}(\frac{(\boldsymbol{x}_{t} - \boldsymbol{\gamma}_{t}\boldsymbol{x}_{t-1})^{2}}{\boldsymbol{\sigma}_{t}^{2}}+\frac{(\boldsymbol{x}_{t-1} - \bar{\boldsymbol{\gamma}}_{t-1}\boldsymbol{x}_{0})^{2}}{\bar{\boldsymbol{\sigma}}_{t-1}^{2}}-\frac{(\boldsymbol{x}_{t} - \bar{\boldsymbol{\gamma}}_{t}\boldsymbol{x}_{0})^{2}}{\bar{\boldsymbol{\sigma}}_{t}^{2}}))\\
=&\exp(((\frac{\boldsymbol{\gamma}_{t}}{\boldsymbol{\sigma}_{t}})^{2}+(\frac{1}{\bar{\boldsymbol{\sigma}}_{t-1}})^{2})\boldsymbol{x}_{t-1}^{2}-(\frac{2\boldsymbol{\gamma}_{t}}{\boldsymbol{\sigma}_{t}^{2}}\boldsymbol{x}_{t}+\frac{2\bar{\boldsymbol{\gamma}}_{t-1}}{\bar{\boldsymbol{\sigma}}_{t-1}^{2}} \boldsymbol{x}_{0})\boldsymbol{x}_{t-1}+C(\boldsymbol{x}_t,\boldsymbol{x}_0)),
\end{aligned}
\end{equation}
in which the recursive relationship is used: \( \bar{\boldsymbol{\sigma}}_{t}^{2} = \boldsymbol{\gamma}_{t}^{2} \bar{\boldsymbol{\sigma}}_{t-1}^{2} + \boldsymbol{\sigma}_{t}^{2} \), which can be inferred by combining \eqn\ref{eqn:forward-t-1-t} and \eqn\ref{eqn:forward-0-t}.

\clearpage